\documentclass[10pt,journal,compsoc]{IEEEtran}
\ifCLASSOPTIONcompsoc
  \usepackage[nocompress]{cite}
  \usepackage{amsmath,amsfonts}
  \usepackage{algorithmicx}
  \usepackage{algpseudocode}
  \usepackage{array}
  \usepackage{amsmath}
  \usepackage{amsmath, bm}
  
  \usepackage{enumitem}
  \usepackage[caption=false,font=normalsize,labelfont=sf,textfont=sf]{subfig}
  \usepackage{booktabs}
  \usepackage{bbding}
  \usepackage{dsfont}
  \usepackage{color} 
  
  \usepackage[ruled,linesnumbered]{algorithm2e}
  \usepackage[pagebackref=true,breaklinks=true,colorlinks=true,bookmarks=false,linkcolor=red,anchorcolor=black,citecolor=green, urlcolor=black]{hyperref}
  \usepackage[table]{xcolor}
  \usepackage{textcomp}
  \usepackage{pifont}
  \usepackage{stfloats}
  \usepackage{enumerate}
  \usepackage{url}
  \usepackage{verbatim}
  \usepackage{graphicx}
  \usepackage{cite}
  \usepackage{bbm}
  \usepackage{multirow}
\else
  \usepackage{cite}
\fi
\ifCLASSINFOpdf
\else
\fi
\begin{document}
%
\title{Prompt-and-Transfer: Dynamic Class-aware Enhancement for Few-shot Segmentation}
%
%
%
%

\author{Hanbo~Bi,~
Yingchao~Feng,~\IEEEmembership{Member,~IEEE,}
Wenhui~Diao,~\IEEEmembership{Member,~IEEE,}
Peijin~Wang,~\IEEEmembership{Member,~IEEE,}\\ 
Yongqiang~Mao,~\IEEEmembership{Member,~IEEE,} 
Kun~Fu,~\IEEEmembership{Member,~IEEE,}
Hongqi~Wang,~\IEEEmembership{Member,~IEEE,}\\
and~Xian~Sun,~\IEEEmembership{Senior Member,~IEEE}
\IEEEcompsocitemizethanks{
\IEEEcompsocthanksitem H. Bi, K. Fu, X. Sun are with the Aerospace Information Research Institute, Chinese Academy of Sciences, Beijing 100190, China, also with the School of Electronic, Electrical and Communication Engineering, University of Chinese Academy of Sciences, Beijing 100190, China, also with the University of Chinese Academy of Sciences, Beijing 100190, China, and also with the Key Laboratory of Network Information System Technology (NIST), Aerospace Information Research Institute, Chinese Academy of Sciences, Beijing 100190, China.
\IEEEcompsocthanksitem Y. Feng, W. Diao, P. Wang, and H. Wang are with the Aerospace Information Research Institute, Chinese Academy of Sciences, Beijing 100190, China, and also with the Key Laboratory of Network Information System Technology (NIST), Aerospace Information Research Institute, Chinese Academy of Sciences, Beijing 100094, China.
\IEEEcompsocthanksitem Y. Mao is with the Department of Electronic Engineering, Tsinghua University, Beijing 100084, China.
\IEEEcompsocthanksitem Corresponding authors: X. Sun (e-mail: sunxian@aircas.ac.cn).

}
}

%
%

\markboth{Journal of \LaTeX\ Class Files,~Vol.~14, No.~8, August~2015}%
{Shell \MakeLowercase{\textit{et al.}}: Bare Demo of IEEEtran.cls for Computer Society Journals}
%



\IEEEtitleabstractindextext{%
\begin{abstract}
For more efficient generalization to unseen domains (classes), most Few-shot Segmentation (FSS) would directly exploit pre-trained encoders and only fine-tune the decoder, especially in the current era of large models. However, such fixed feature encoders tend to be class-agnostic, inevitably activating objects that are irrelevant to the target class. In contrast, humans can effortlessly focus on specific objects in the line of sight. This paper mimics the visual perception pattern of human beings and proposes a novel and powerful prompt-driven scheme, called ``Prompt and Transfer" (PAT), which constructs a dynamic class-aware prompting paradigm to tune the encoder for focusing on the interested object (target class) in the current task. Three key points are elaborated to enhance the prompting: 1) Cross-modal linguistic information is introduced to initialize prompts for each task. 2) Semantic Prompt Transfer (SPT) that precisely transfers the class-specific semantics within the images to prompts. 3) Part Mask Generator (PMG) that works in conjunction with SPT to adaptively generate different but complementary part prompts for different individuals. Surprisingly, PAT achieves competitive performance on 4 different tasks including standard FSS, Cross-domain FSS (e.g., CV, medical, and remote sensing domains), Weak-label FSS, and Zero-shot Segmentation, setting new state-of-the-arts on 11 benchmarks.

\end{abstract}

\begin{IEEEkeywords}
Few-shot Learning, Semantic Segmentation, Few-shot Segmentation, Prompt Learning
\end{IEEEkeywords}}

\maketitle

\IEEEdisplaynontitleabstractindextext

%
\IEEEpeerreviewmaketitle

\IEEEraisesectionheading{\section{Introduction}\label{sec:introduction}}
\IEEEPARstart{W}{ith} the growing wave of deep learning, various computer vision tasks have made remarkable progress, such as image classification, object detection, semantic segmentation, spatiotemporal prediction, and 3D reconstruction \cite{he2016identity,he2016deep,girshick2015fast,redmon2016you,lin2017feature,long2015fully,ronneberger2015u,10254320,10440371,mao2022beyond,mao2023elevation,10522786}. However, such techniques are data-driven and tend to fail with satisfactory performance when labeled data is insufficient \cite{reichstein2019deep,zhuang2020comprehensive}. Even though semi-supervised learning has been suggested to tackle the problem of data scarcity, they fail to generalize well to the unseen (novel) classes \cite{berthelot2019mixmatch,zhai2019s4l,van2020survey}. In contrast, humans can recognize new patterns or concepts easily from a few examples. The vast gap has tremendously aroused the interest of researchers. Thus, Few-shot Learning (FSL) has been developed to generalize quickly to unseen domains (classes) with a handful of labeled samples \cite{finn2017model,vinyals2016matching,ravi2016optimization,xu2024attention}. 

\begin{figure*}[t]
\centering
\setlength{\abovecaptionskip}{1pt}
\setlength{\belowcaptionskip}{1pt}
\includegraphics[width=0.95\linewidth]{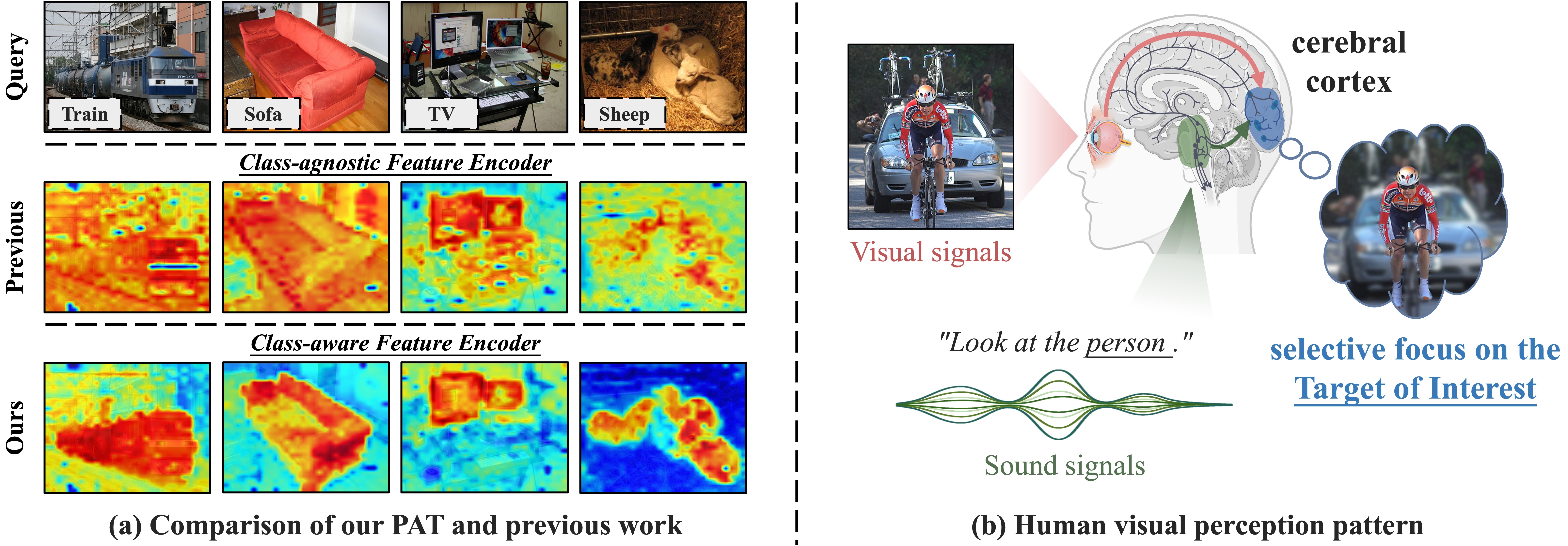}
\caption{ (a) Comparison of our PAT and previous work. For more efficient generalization, most FSS methods prefer to directly employ the pre-trained encoders and only fine-tune the decoder. However, such frozen feature encoders tend to be class-agnostic, inevitably activating other classes irrelevant to the current FSS task (the 2nd row), due to their semantic clues derived from pre-training on image classification. (b) Human visual perception pattern. When processing visual stimuli from the retina, the cerebral cortex simultaneously parses stimuli from the mental, sound, etc., in the current state, selectively focusing on specific objects in view while leaving the rest parts in the shadow of consciousness. Inspired by the unique visual perception pattern, our ``Prompt and Transfer" (PAT) method instead dynamically drives the encoder to focus on specific objects in a class-aware prompting manner (the 3rd row in (a)).}
\label{fig:1}
\end{figure*}

Few-shot Segmentation (FSS) is the application of FSL to the task of semantic segmentation \cite{shaban2017one}. Given the original (query) image, the FSS model aims at segmenting objects of the specific category in the query image by utilizing one or several labeled samples, namely support images. Remarkably, the categories segmented are unseen and rely solely on several support images to provide target category information. For more efficient generalization to unseen classes, most FSS researchers would directly employ pre-trained encoders (e.g., ResNet-50 trained on ImageNet-1K \cite{russakovsky2015imagenet}) and only fine-tune the decoder \cite{zhang2020sg,nguyen2019feature,wang2019panet,liu2020part,tian2020prior,chen2021apanet,min2021hypercorrelation}. A major reason for this trend is to prevent the feature encoder from being biased towards seen (base) classes during the training phase \cite{lang2024few,liu2022learning,mao2022bidirectional}.

\begin{figure*}[t]
\centering
\setlength{\abovecaptionskip}{1pt}
\setlength{\belowcaptionskip}{1pt}
\includegraphics[width=0.95\linewidth]{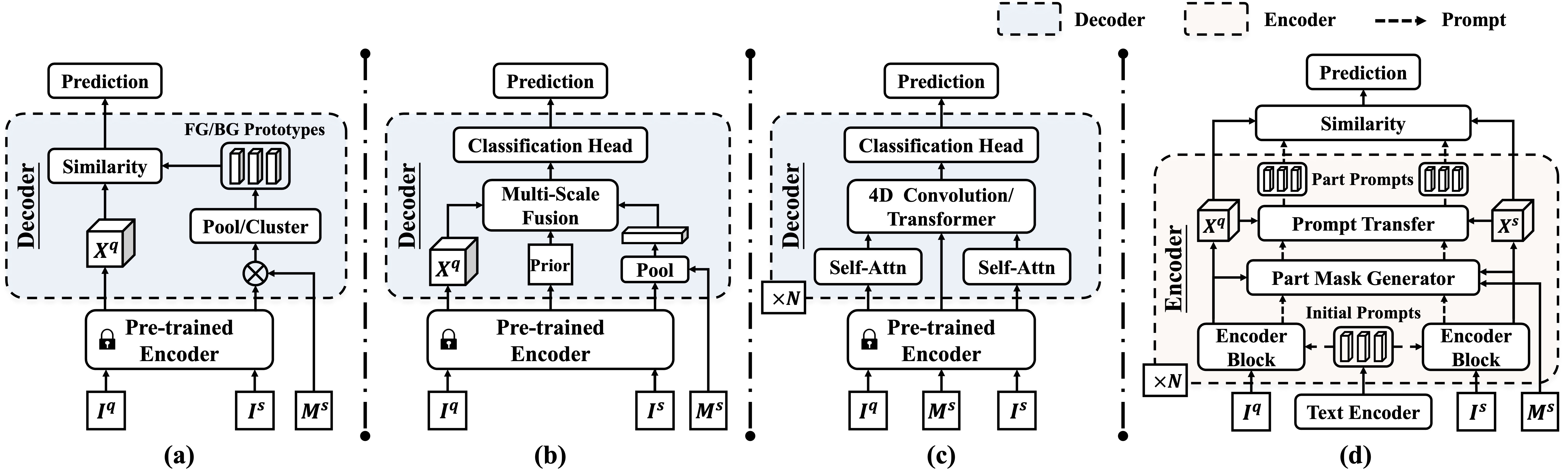}
\caption{ Overview of various Few-shot Segmentation (FSS) structures. (a) Prototype Matching-based methods \cite{zhang2020sg,wang2019panet,liu2020part,fan2022self}. (b) Feature Fusion-based methods \cite{chen2021apanet,min2021hypercorrelation,cheng2022holistic,liu2022learning}. (c) Pixel Matching-based methods \cite{zhang2021few,wang2022adaptive,xu2023self,peng2023hierarchical}.  (d) Our "Prompt and Transfer" (PAT) method, dynamically generates part-level semantic prompts (i.e. part prompts) to tune the encoder for activating class-specific objects in the query image.}
\label{fig:2}
\end{figure*}

Despite the promising progress, we observe that these methods overlook a critical issue, i.e., such feature encoders with fixed parameters tend to be class-agnostic. In other words, they will inevitably activate other classes irrelevant to the target class in FSS tasks, due to their semantic clues derived from pre-training on image classification (e.g., ImageNet-1K). As depicted in Fig.\ref{fig:1}(a), the encoder fails to effectively focus on task objects such as "Train" and "Sofa", and likewise activates irrelevant regions around them. Such cluttered features would exacerbate the burden on the subsequent decoder to segment novel classes \cite{min2021hypercorrelation,wang2022adaptive,yang2020prototype,bi2023not}. Even though FSS researchers have designed various powerful decoders such as HSNet's 4D convolution~\cite{min2021hypercorrelation} and CyCTR's transformer variant~\cite{zhang2021few}, to perform more precise feature matching for superior segmentation \cite{shaban2017one,boudiaf2021few,yang2021mining,lang2022learning,chen2021apanet,liu2022intermediate,fan2022self,bao2023relevant,lang2023progressive} (see Fig.\ref{fig:2}(a)-(c)), the irrelevant activation introduced by the frozen encoder has not been substantially addressed.

Humans can selectively focus on critical objects in a unique pattern of visual perception, which is because "\textit{we are not passive recipients of retinal information, but active participants in the perceptual process}" \cite{kanwisher2000visual}. As depicted in Fig.\ref{fig:1}(b)), when processing stimuli from the retina, the cerebral cortex simultaneously parses other stimuli from the psyche, sound, etc., selectively focusing on specific objects in view while leaving other parts in the shadow of consciousness. Inspired by this, we argue that the ideal feature encoder in FSS should be class-aware, activating corresponding class objects for different tasks, even for unseen classes. Prompt learning \cite{brown2020language,li2021prefix,jia2022visual,zhou2022learning}, which aims to construct several prompt words or vectors to provide prior task information to the model and adapt the model behavior to task-specific patterns, provides us with the solution. Thus, this work proposes a prompt-driven scheme for FSS, dynamically prompting the encoder to focus on class-specific objects in the query image for class-aware enhancement (see Fig.\ref{fig:2}(d)), called "Prompt and Transfer", i.e., \textbf{PAT}.

Concretely, the core of PAT is to adaptively learn class-specific prompts from the support- and query- pixels in the current task, which will continuously interact with image features through the self-attention structure \cite{touvron2021training,dosovitskiy2020image} of the encoder to generate the image features focused on specific objects, i.e., class-aware features. In this case, without the need to design complex decoders, effective segmentation can be achieved by utilizing these prompts for simple similarity computation with the class-aware features. Against this background, it is crucial to dynamically derive the corresponding effective prompts for different tasks. PAT elaborates on three key points for the enhancement of dynamic class-aware prompting:

\textbf{Firstly,} we introduce cross-modal linguistic information to endow prompts with initial class-awareness in each task. The insight behind such an operation is that the pre-trained language model can characterize category-representative textual semantics based on different task descriptions, thus initially localizing the target object in the visual feature. \textbf{Secondly,} we design the Semantic Prompt Transfer (SPT) to continuously enhance the dynamic class-awareness of prompts. For better tuning the encoder to focus on the target class in the current task, SPT then adaptively transfers the target semantics within a specific region (e.g., foreground region) to the prompts via the cross-attention manner, where Gaussian suppression is proposed to optimize the attention distribution. Considering the vast gap between the image pair resulting from the large intra-class variations and cluttered background, singularly capturing target semantics from the support image is inadequate. We attempt to construct the pseudo-mask for the query image and then utilize it for guiding SPT to capture the target clues of the query itself, thus bridging the semantic gap. \textbf{Thirdly,} we construct the Part Mask Generator (PMG) to mine fine-grained part-level semantic prompts. To enhance the semantic diversity of the prompts, PMG adaptively generates a series of part-level local masks for different individuals in different tasks. The SPT, with the help of PMG, can constrain different prompts to aggregate the semantics under different part masks, yielding different but complementary part prompts. In this way, the redundancy of prompts is avoided and local semantic clues are fully explored simultaneously. 

\begin{figure}[t]
\setlength{\abovecaptionskip}{1pt}
\setlength{\belowcaptionskip}{1pt}
\centering
\includegraphics[width=0.95\linewidth]{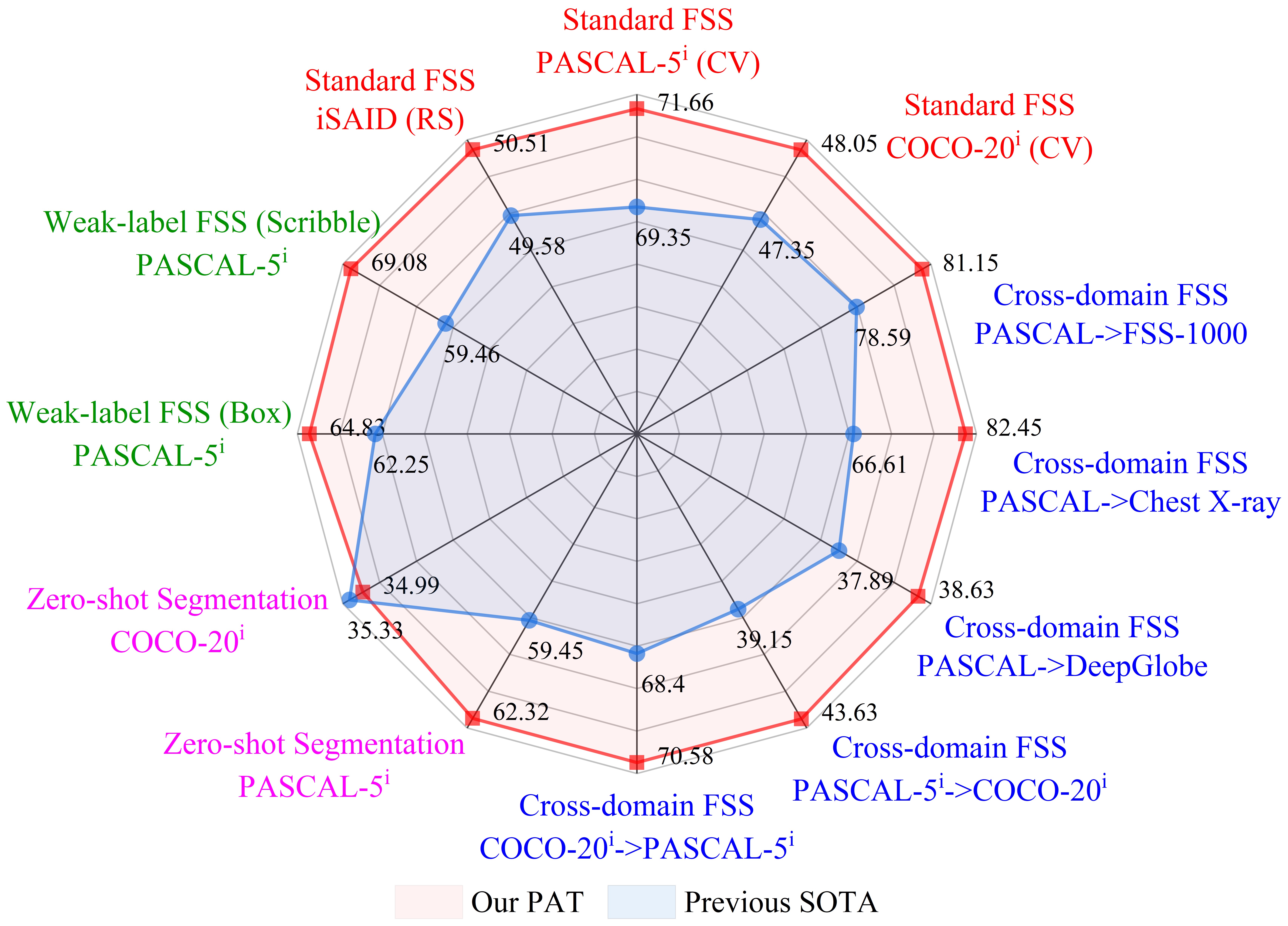}
\caption{Our PAT has performed excellently (11 received SOTA) on 12 benchmarks over 4 different tasks. All results are obtained with the backbone of Deit-B/16 under the 1-shot setting (except for Zero-shot Segmentation).}
\label{fig:0}
\end{figure}

Extensive experiments on popular FSS benchmarks validate the effectiveness of PAT, which sets new state-of-the-art performance. Moreover, due to its flexibility, we extend it to three more realistic yet complex scenarios, including Cross-domain FSS (e.g., CV, medical, and remote sensing domains), Weak-label FSS, and Zero-shot Segmentation (ZSS) (see Fig.\ref{fig:0}). In summary, our primary contributions can be summarized as follows:
\begin{itemize}
    \item This paper mimics the visual perception pattern of humans and develops a novel dynamic class-aware prompting paradigm (PAT) to tune the encoder for focusing on specific objects in different FSS tasks. 

    \item Three key points including Prompt Initialization, Semantic Prompt Transfer, and Part Mask Generator, are constructed to assign class-specific semantics to prompts for better enhancing the dynamic class-aware prompting.

    \item The prompt-driven scheme establishes new state-of-the-arts under all settings on three standard FSS benchmarks, including PASCAL-5$^i$, COCO-20$^i$ and iSAID.
    
    \item We extend the proposed method to three more realistic yet challenging tasks, including Cross-domain, Weak-label, and Zero-shot Segmentation, comprehensively demonstrating its versatility and flexibility.

\end{itemize}

\section{Related Work} 
\subsection{Few-shot Learning}
Recently, community researchers have proposed Few-shot Learning (FSL) to explore the generalizability in unseen domains, which can quickly identify unseen classes with limited samples. The vast majority of current methods follow the meta-learning paradigm \cite{finn2017model}, where transferable meta-knowledge is learned from a series of tasks (i.e., episodes) sampled from a base dataset and generalized to novel tasks. On this basis, such methods can be further subdivided into two branches: (1) Optimization-based methods \cite{finn2017model,li2017meta,ravi2016optimization,nichol2018first}, aiming to derive the most appropriate parameters to enable the model to be rapidly adapted to novel tasks. (2) Metrics-based methods \cite{vinyals2016matching,snell2017prototypical,sung2018learning}, aiming to learn a specific distance metric, e.g., cosine similarity distance \cite{vinyals2016matching}, Euclidean distance\cite{snell2017prototypical}, to derive similarities between different objects in the embedding space. Inspired by the utilization of textual information for classification in the recent FSL method \cite{chen2023semantic}, we attempt to introduce the textual information of language modality to the segmentation task for tackling the lack of semantics under the low-data regime. 

\subsection{Few-shot Segmentation} \label{sec:Few-shot Segmentation}
Few-shot Segmentation (FSS) is the extension of Few-shot Learning for handling dense prediction tasks that aim to segment unseen classes in the query image with several labeled samples \cite{shaban2017one}. 
Previous FSS methods can be divided broadly into three branches: (1) prototype matching-based methods (see Fig.\ref{fig:2}(a)) \cite{zhang2020sg,wang2019panet,liu2020part,fan2022self,lang2024few}, which compress the support features into one or multiple category-representative prototypes through Mask Average Pooling \cite{zhang2020sg}. For example, SSPNet~\cite{fan2022self} aims to aggregate prototypes from semantic clues of the query image itself to guide its segmentation. DCPNet~\cite{lang2024few} subdivides the support features into several prototype proxies with different properties. (2) feature fusion-based methods (see Fig.\ref{fig:2}(b)) \cite{chen2021apanet,min2021hypercorrelation,cheng2022holistic,liu2022learning,yang2023mianet,lang2023retain}, which aggregates support- and query- features for feature fusion. For example, HSNet~\cite{min2021hypercorrelation} constructs multi-scale feature fusion utilizing 4D convolution. IPMT~\cite{liu2022intermediate} integrates support- and query- semantics into an intermediate prototype to bridge the gap between image pairs. (3) pixel matching-based methods (see Fig.\ref{fig:2}(c)) \cite{zhang2021few,wang2022adaptive,xu2023self,peng2023hierarchical}, which further consider the similarity between support-query pixel pairs. For example, CyCTR~\cite{zhang2021few} aggregates the pixel-level semantics of the support image via the cross-attention structure. Overall, the majority of previous methods simply employ the frozen pre-trained encoders and focus on designing various decoders to achieve superior feature matching.

Recently several researchers have indicated the potential and necessity of exploring a feature encoder that serves the FSS task, in addition to designing powerful decoders. SVF \cite{sun2022singular} argues that fine-tuning a small part of the encoder parameters not only avoids overfitting the base class but also allows for a better generalization to novel classes. Specifically, it decomposes the parameters of the encoder into three successive matrices via the Singular Value Decomposition (SVD), where only the singular value matrix is fine-tuned while keeping others frozen. In addition, BAM \cite{lang2022learning} additionally trains a base learner (i.e., encoder) to explicitly identify base classes, i.e., regions that do not need to be segmented, to better focus on novel classes. Different from them, this work develops a novel dynamic class-aware encoder that selectively focuses on corresponding classes within the image for different tasks (i.e., novel classes) in a prompting manner.

\subsection{Prompt Learning} \label{sec:Prompt Learning}
Prompt Learning was first proposed in NLP \cite{brown2020language,li2021prefix,gao2020making,liu2023pre}, which aims to construct several prompt words or vectors to provide prior task information to the model and adapt the model behavior to task-specific patterns. Inspired by its huge success, some recent works in CV have attempted to introduce a series of learnable parameters to activate the corresponding semantic knowledge in the visual model for different tasks \cite{jia2022visual,chen2023semantic,zhou2022learning}. Such the paradigm of prompting models to activate different semantics for different tasks matches well with the meta-learning framework in FSL, which expects to learn generalizability from a series of tasks. Recent works \cite{tsimpoukelli2021multimodal,zhang2022feature} have explored combining prompt learning and few-shot learning. For example, SP~\cite{chen2023semantic} leverages semantic information as prompts to tune the visual feature encoder adaptively in the Few-shot Classification task. FPTrans \cite{zhang2022feature} directly takes the support prototypes as prompts, which interact with the query feature in the encoder to activate the object. However, forcing the support prototypes as prompts to activate the query feature may be sub-optimal due to the differences between image pairs. In this paper, we construct class-specific prompts for different tasks (i.e., novel classes). For further enhancing prompting, besides introducing cross-modal linguistic information to initialize prompts, Semantic Prompt Transfer and Part Mask Generator are proposed to transfer the fine-grained local semantics of support and query images to prompts.

\begin{figure*}[t]
\centering
\setlength{\abovecaptionskip}{1pt}
\setlength{\belowcaptionskip}{1pt}
\includegraphics[width=0.92\linewidth]{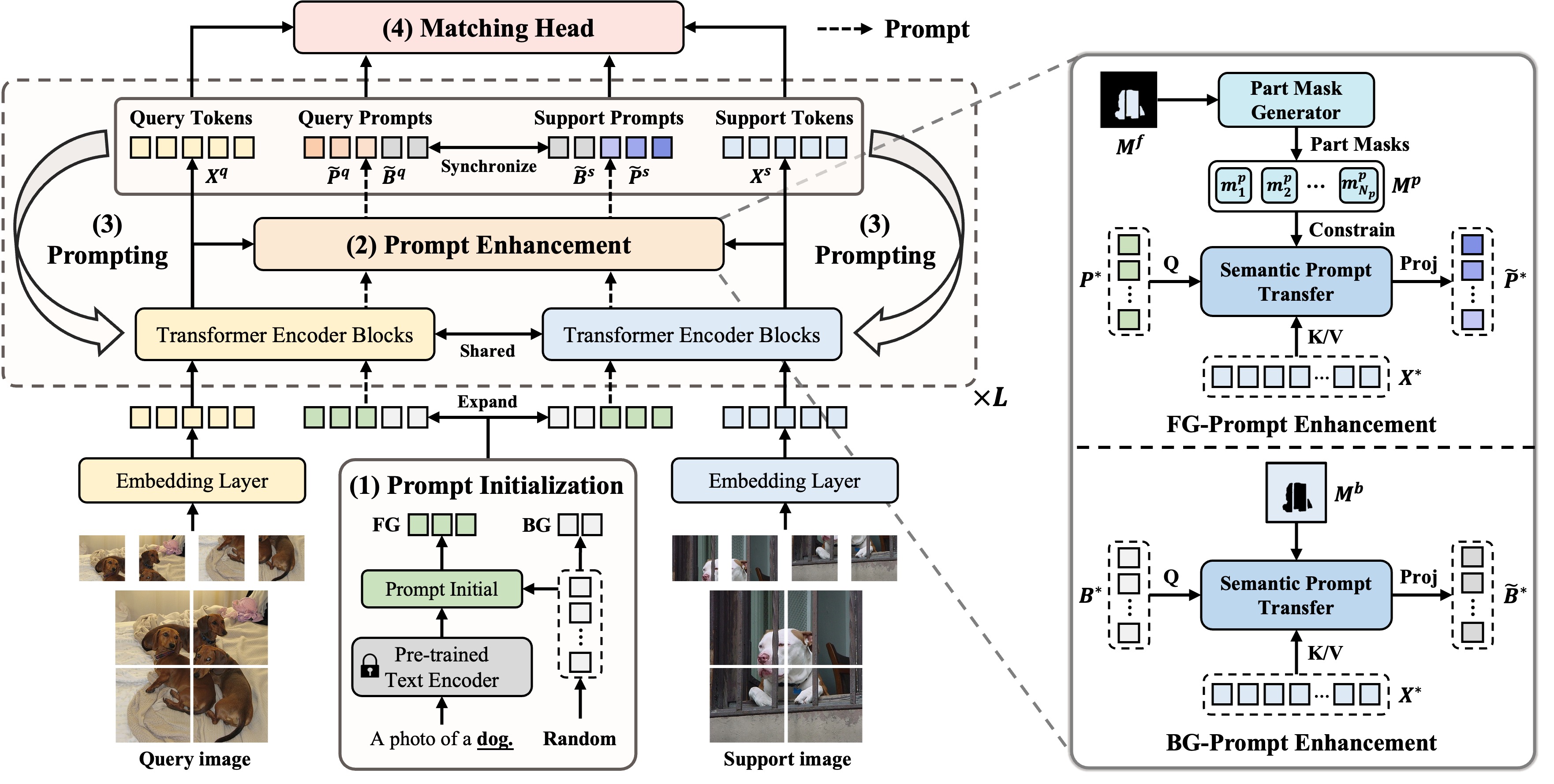} 
\caption{Overview structure of the proposed PAT. PAT first derives the image tokens through the Embedding layer. Then a pre-trained text encoder is introduced to mine representative textual semantics, which along with randomly initialized embeddings are utilized as the initial prompts to interact with image features. For better enhancing the dynamic class-aware prompting, Prompt Enhancement is introduced to adaptively transfer the target semantics within a specific region (e.g., fine-grained local regions) from the support/query image to prompts via the Semantic Prompt Transfer and Part Mask Generator. These prompts will in turn interact with image features in the next encoder block to activate specific objects within the features. After several alternations of prompting and transferring in the encoder blocks, the derived prompts are directly utilized to perform similarity computation with the class-aware query feature to produce the segmentation results in the Matching Head. Notably, we only describe the encoder blocks that perform the Prompt Enhancement, while omitting others.}
\label{fig:3}
\end{figure*}

\section{Methodology}
\subsection{Problem Definition}
\textbf{Standard FSS:} The purpose of Few-shot Segmentation (FSS) is to segment unseen objects utilizing several labeled images. Specifically, given two datasets \emph{D}$_{train}$ and \emph{D}$_{val}$ with seen classes \emph{C}$_{seen}$ and unseen classes \emph{C}$_{unseen}$ ($ C_{seen} \cap C_{unseen}= \emptyset$), respectively, the FSS model is trained on \emph{D}$_{train}$ and generalized directly to \emph{D}$_{val}$. Following the previous methods \cite{shaban2017one,wang2019panet}, episode training is employed in this work: Each episode will contain a support set $ S =\left \{ \left ( I_i^s, M_i^s\right )  \right \}_{i=1}^{K} $  and a query set $ Q =\left \{ \left ( I^q, M^q\right )  \right \} $, where $I^{*}_{i}$ and $M^{*}_{i}$ denote the image and the corresponding mask. Notably, both $I^{s}$ and $I^{q}$ contain the same category and $K$ indicates the number of labeled samples provided in the support set, where we just discuss the segmentation performance under the 1-shot and 5-shot settings, i.e., $K=1,5$. In every episode, the model will segment the specific objects in the query image $I^{q}$ guided by the support set \emph{S}.

Variant: In previous work, the masks $M_i^s$ are usually represented as one-hot vectors, e.g., $M_i^s=[1,0,1,0,...]$. However, such numerical representations ignore the valuable linguistic semantics of the class labels (e.g., "dog", "car"). This work additionally introduces the class label $M^t$ to mine the text semantics, at which point the support set $ S =\left \{ \left ( I_i^s, M_i^s, M^t \right )  \right \}_{i=1}^{K} $.

\noindent \textbf{Cross-domain FSS:} The samples during the training and testing phases are from different domains (datasets), rather than the same domain \cite{lu2021simpler,lei2022cross}.

\noindent \textbf{Weak-label FSS:} The Support images with sparse annotations, e.g., scribbles, bounding boxes, rather than dense annotations, are provided to guide segmentation \cite{wang2019panet}.

\noindent \textbf{Zero-shot Segmentation:} Only the task (category) information is provided to guide segmentation without any annotated samples (support images) \cite{xian2019semantic,bucher2019zero}.

\subsection{Method Overall}
Recent works \cite{hassabis2017neuroscience,du2021ago} have proven that exploiting the biological plausibility of the human brain to build up intelligent systems brings significant performance gains for various tasks. With a unique pattern of visual perception, humans can selectively focus on critical objects in the line of sight while leaving the rest in the shadow of consciousness \cite{kanwisher2000visual}. This motivates us to explore the dynamic class-aware feature encoder, which can adaptively activate corresponding class objects for different tasks. Based on the prompt learning technology\cite{jia2022visual,chen2023semantic,zhou2022learning}, we expect to construct dynamic semantic prompts to adjust the encoder in FSS.

Thus, we propose introducing a dynamic class-aware prompting paradigm to tune the encoder for focusing on the class-specific objects in the current task. Based on this prompting pattern, we construct a novel prompt-driven FSS framework called "Prompt and Transfer", i.e., PAT. Fig.\ref{fig:3} illustrates the overall architecture of PAT. Given the support image $I^s$ and query image $I^q$, we feed them into the patch embedding layer to derive support tokens $X^s \in \mathbb{R}{^{N_l \times d}}$ and query tokens $X^q \in \mathbb{R}{^{N_l \times d}}$ ($N_l$ embedding vectors with $d$ dimensions, unless otherwise noted, support and query tokens are collectively denoted as $X$). In summary, our PAT consists of the following four key steps:

\textbf{1) Prompt Initialization:} PAT introduces a typical pre-trained visual language model to extract the text semantics in category names and add them to several randomly initialized embeddings exploited as initial prompts.

\textbf{2) Prompt Enhancement:}
The Prompt Enhancement consists of a Semantic Prompt Transfer (SPT) and a Part Mask Generator (PMG), which transfers semantics within specific regions (e.g., fine-grained local regions) of the support/query images to prompts, enhancing class-awareness of prompts for adjusting the encoder.

\textbf{3) Prompting:} After transferring, such prompts will in turn interact with the image features in the next encoder block. This alternation of prompting and transferring can prompt the model to further activate the objects, yielding more discriminative class-aware features for every task.

\textbf{4) Matching:} In this case, effective segmentation can be achieved by simply utilizing these prompts for similarity computation with the enhanced query feature.

\subsection{Prompt Initialization} \label{Prompt Initialization}
The prompting process is implemented via the self-attention of the transformer \cite{jia2022visual}. Through the interaction of prompts and image features in the self-attention structure, images can activate specific target regions while prompts further aggregate target semantics from images. At this time, the source of the prompt will greatly affect its performance. As mentioned in Sec.\ref{sec:Prompt Learning}, choosing the support mask feature as semantic prompts is not optimal due to the information differences between the image pair. We argue that other modal semantics might be a suitable source. Given that textual information being strongly representative can nicely characterize the category semantics of different tasks, we utilize it as the initial prompt to tune the image feature. 

Concretely, as illustrated in Fig.\ref{fig:3}, the Pre-trained visual language model CLIP \cite{radford2021learning} (frozen during the whole process) is employed to extract the text feature from the uniform template ``\texttt{a photo of a [CLS]}.'' where [CLS] is the class label $M^t$. The two-layer MLP module is then utilized to align it with the visual features. At this point, it is feasible to directly utilize this text embedding as an initial prompt. To enhance the effectiveness of prompting, following the previous prompt learning methods~\cite{jia2022visual,zhou2022learning,chen2023semantic}, we employ multiple prompts to tune the encoder. More specifically, we utilize the Xavier uniform initialization scheme to generate multiple learnable embeddings $t_i$ with mean 0, then sum (add) these learnable embeddings with the text embedding generated by CLIP to derive our initial FG-prompts (foreground prompts), i.e., $P = \left [ p_1,p_2,...,p_{N_p} \right ] \in \mathbb{R}{^{N_p \times d}}$, which is equivalent to sampling multiple learnable embeddings from a normal distribution with mean text embedding:
\begin{equation}\label{eq:1}
\setlength{\abovecaptionskip}{1pt}
\setlength{\belowcaptionskip}{1pt}
p_i = {\mathcal{F}}_{M}\left ({\mathcal{F}}_{T}\left (M^t\right ) \right )+t_i, i=1,2,...,N_p 
\end{equation}
where ${\mathcal{F}}_{T}$ indicates the text encoder in CLIP, ${\mathcal{F}}_{M}$ indicates the MLP module, $M^t$ indicates the class label in every episode, $N_p$ indicates the number of FG-prompts. In addition, giving the importance of background information for segmentation, we randomly initialize $N_b$ embeddings as the initial BG-prompts (background prompts), i.e., $B = \left [ b_1,b_2,...,b_{N_b} \right ] \in \mathbb{R}{^{N_b \times d}}$, which adaptively learn the background semantics in the subsequent process.

Thus, the prompts concatenated with feature tokens are fed into the transformer together for interactions.

\subsection{Prompt Enhancement} \label{Prompt Enhancement}
For better enhancing the dynamic class-aware prompting, we suggest capturing richer class-specific semantics from the support and query images. Thus, the Prompt Enhancement is performed to transfer target semantics from the support/query feature to the proposed prompts, which consists of the FG-Prompt Enhancement and the BG-Prompt Enhancement (see Fig.\ref{fig:3}).

Take the FG-Prompt Enhancement as an example, the whole enhancement is implemented via two main components: a Semantic Prompt Transfer (SPT) and a Part Mask Generator (PMG). The former is the core, which constructs robust semantic transfer between feature tokens and prompts under the constraints of the specific mask (e.g., foreground) via a cross-attention manner, in which we design Gaussian suppression to avoid preferring the few most discriminative pixels while neglecting other valuable pixel semantics. To prevent all prompts from aggregating similar category semantics during the migration, we construct a Part Mask Generator (PMG) that adaptively generates a series of different part-level masks for different individuals. With the help of PMG, Semantic Prompt Transfer (SPT) can constrain different prompts to aggregate the semantics under different part masks, yielding different but complementary FG-prompts (part prompts). In this way, the redundancy of prompts is avoided and the local semantic clues are fully explored at the same time. The details are shown in Algorithm \ref{FG-Prompt Enhancement}.

\begin{algorithm}
\setlength{\abovecaptionskip}{1pt}
\setlength{\belowcaptionskip}{1pt}
\small
\caption{ FG-Prompt Enhancement }\label{FG-Prompt Enhancement}
\KwIn{ image feature $X$, FG-prompts $P$}
\KwOut{ Class-aware FG-prompts $\widetilde{P}$}

\textbf{Part Mask Generator} (Sec.\ref{Part Mask Generator})

First, generate the parameters of Part Filters $\varphi \leftarrow P$\ according to Eq.(\ref{eq:2});

\For{$\varphi_i$ in $\varphi$ }{
Derive the part mask by the Part Filter $M^p_i \leftarrow \left (\varphi_i, X\right )$\ according to Eq.(\ref{eq:3});
}
\textbf{Semantic Prompt Transfer} (Sec.\ref{Semantic Prompt Transfer})

\For{$P_i$ in $P$ }{
Transfer the semantics of the $i^{th}$ part mask to the prompt
$\widetilde{P_i} \leftarrow \left (P_i, X, M^p_i\right )$\ according to Eq.(\ref{eq:6});
}
\end{algorithm}

\subsubsection{ Part Mask Generator (PMG)} \label{Part Mask Generator} 
Part Mask Generator (PMG) aims to adaptively generate different but complementary part masks for different individuals (images). As illustrated in Fig.\ref{fig:4}, multiple Part Filters, i.e., $\varphi=\left \{ \varphi_1,\varphi_2,...,\varphi_{N_p} \right \} $, are introduced to generate part-level local masks in PMG, with different filters responsible for filtering different part regions. However, since different tasks need to focus on different category objects, it is sub-optimal to share the parameters with the same set of Part Filters across different tasks. In this case, we introduce a Part Filter Learner, i.e., $G_p = \left \{ g_1,g_2,...,g_{N_p} \right \} $, to adaptively learn filter parameters for different tasks. Concretely, the learner $G_p$ takes the current FG-prompts (from support or query) as input and maps the class-specific semantics contained in the prompts as the parameters of different Part Filters, which can be formulated as:
\begin{equation}\label{eq:2}
\setlength{\abovecaptionskip}{1pt}
\setlength{\belowcaptionskip}{1pt}
\varphi_i = g_i \left( p_i \right),i=1,2,...,N_p
\end{equation}
where $g_i $ is formed by a two-layer MLP module and $p_i$ denotes the $i^{th}$ FG-prompt. With this learner, the Part Filter can adaptively obtain the most suitable set of parameters for different tasks (classes), thus generating part regions more accurately.

After that, we apply the Part Filters $\varphi \in \mathbb{R}{^{ N_p \times d \times1 \times1}}$ as $1\times1$ convolution kernels to map the image features $X$ (reshaped into $H \times W\times d$) into the part activation maps. We then normalize these activation maps for distinction and multiply them by the foreground mask to force attention to the foreground region. Thus, the part masks $M^p= \left \{ m^p_1, m^p_2,..., m^p_{N_p} \right \}  \in \mathbb{R}{^{ N_p \times H \times W }}$ are derived:

\begin{equation}\label{eq:3}
\setlength{\abovecaptionskip}{1pt}
\setlength{\belowcaptionskip}{1pt}
 M^p = \mathrm{softmax}\left ( \bigcup_{i=1}^{N_p} \mathcal{F}_{\varphi_i}\left( X\right ) \right) \odot M^f
\end{equation}

\noindent where $\mathcal{F}_{\varphi}$ denotes the $1\times1$ convolution operation with part filter parameters, $\bigcup$ denotes the concatenation of multiple part activation maps, $\mathrm{softmax}()$ is performed along the dimension of the number of part activation maps, $X$ denotes the query/support feature, $\odot$ represents the dot product operation and $M^f \in \mathbb{R}{^{H \times W }}$ represents the foreground mask of the support/query image. 

Remarkably, we construct a pseudo-mask for the query image, which is derived by weighted aggregation of the support mask with the affinity map between the support- and the query- feature:
\begin{equation}\label{eq:4}
\setlength{\abovecaptionskip}{1pt}
\setlength{\belowcaptionskip}{1pt}
\hat{M^q}=\mathrm{sigmoid} \left ( \frac{X_qX_s^T}{\left \|X_q  \right \| \left \|X_s  \right \|} M^s \right ) 
\end{equation}
where $\hat{M^q}$ is reshaped into $H \times W$ for part mask generation in Eq.(\ref{eq:3}). We expect to exploit this coarse query foreground mask to capture the semantics from the query image itself, instead of utilizing support semantics solely. In this way, performance degradation due to the differences between image pairs can be minimized.

With the Part Mask Generator (PMG), we can obtain the part masks for the support and the query image, respectively, which will be fed to the Semantic Prompt Transfer (SPT) to constrain the route of transferring semantics.

\begin{figure}[t]
\centering
\setlength{\abovecaptionskip}{1pt}
\setlength{\belowcaptionskip}{1pt}
\includegraphics[width=0.85\linewidth]{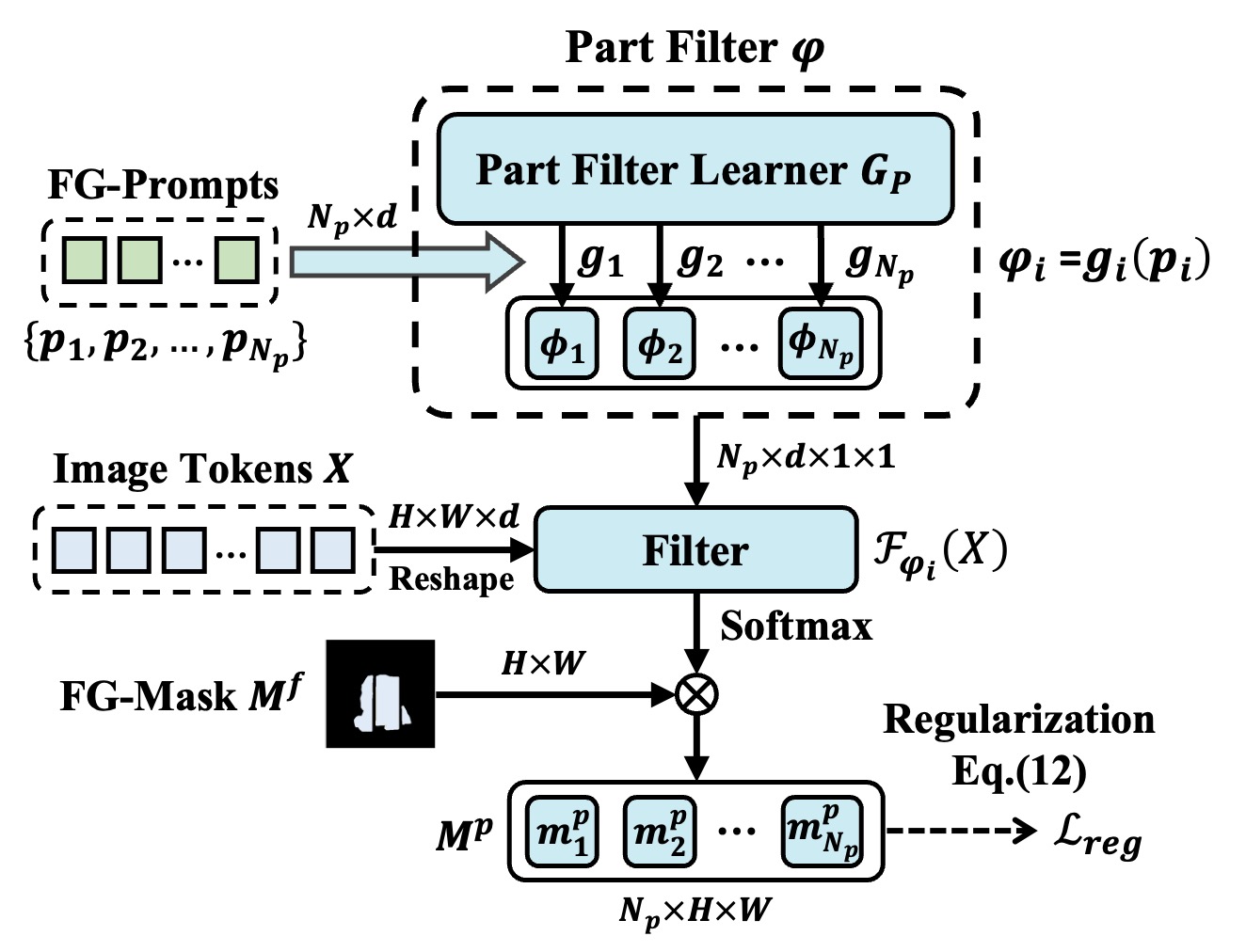} 
\caption{The specific process of the Part Mask Generator (PMG), which aims at adaptively generating a series of different part-level masks for different individuals. }
\label{fig:4}
\end{figure}

\subsubsection{Semantic Prompt Transfer (SPT)} \label{Semantic Prompt Transfer} 
Prompt Enhancement aims at transferring target semantics to prompts for enhancing class-awareness, where exploiting cross-attention is a natural choice. Based on this, we construct a novel Semantic Prompt Transfer (SPT). As illustrated in Fig.\ref{fig:5}, our improvements can be summarized in two parts:

1) To precisely transfer the semantics within a specific region (e.g., the different part mask from PMG) to the prompt rather than the entire feature semantics, the attention region for cross-attention needs to be restricted. Notably, our part masks $M^p\in \left [ 0,1 \right ] $ are probability activation maps not binary masks, which prevents the direct utilization of Masked Attention \cite{cheng2022masked} to adjust attention weights. Thus, we propose to construct a Log-Transformation to map the part masks $M^p$ to the biases of attention weights:
\begin{equation}
\setlength{\abovecaptionskip}{1pt}
\setlength{\belowcaptionskip}{1pt}
\widetilde{m}^p_i = \mathcal{F}_{log}{\left ( m^p_i+\epsilon \right )}, i=1,2,...,N_p 
\end{equation}
where $\mathcal{F}_{log}$ denotes the log-transformation with base 10, with lower mask values indicating higher weight attenuation, and $\epsilon$ is set to $1e-7$ to prevent computational crashes. Via this mapping, the part masks $\widetilde{M}^p=\left \{ \widetilde{m}^p_1, \widetilde{m}^p_2,..., \widetilde{m}^p_{N_p} \right \}  \in \mathbb{R}{^{ N_p \times N_l }} $ will be synchronized into the attention distribution space.

Thus, the attention weight $\widetilde{A}\left (P,X\right) \in \mathbb{R}{^{ N_p \times N_l }}$ can be efficiently constrained to the region of the part masks $M^p$, which can be calculated by:
\begin{equation}
\setlength{\abovecaptionskip}{1pt}
\setlength{\belowcaptionskip}{1pt}
\widetilde{A}_i \left (P,X\right) = A\left (P,X\right) + \widetilde{m}^p_i,i=1,2,...,N_p 
\end{equation}
where $P$ and $X$ represent FG-prompts and image features, respectively. $A\left (P,X\right) = \frac{\left ( PW_q \right ) \left ( XW_k \right ) ^T}{\sqrt{d}} \in \mathbb{R}{^{ N_p \times N_l }}$ denotes the original attention weight ($W_q$ and $W_k$ denote the projection weights). Notably, $\widetilde{A}_i\left (P,X\right) \in \mathbb{R}{^{1 \times N_l }}$ focuses only on the region of $i^{th}$ part mask $m^p_i$.

\begin{figure}[t]
\centering
\setlength{\abovecaptionskip}{1pt}
\setlength{\belowcaptionskip}{1pt}
\includegraphics[width=0.95\linewidth]{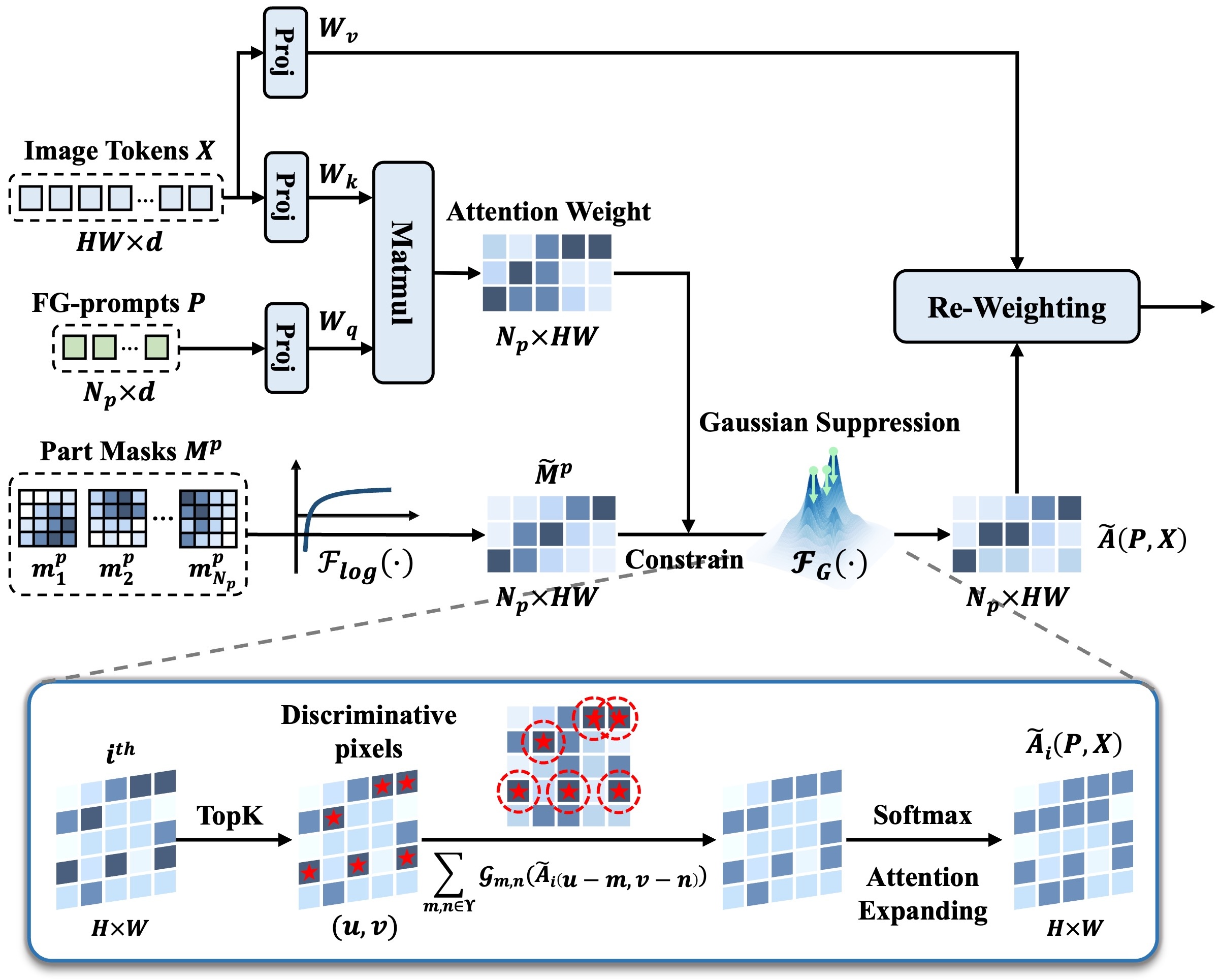} 
\caption{The specific process of the Semantic Prompt Transfer (SPT). Notably, for the BG-Prompt Enhancement, just adjusting the background mask fed into the SPT without the PMG.}
\label{fig:5}
\end{figure}
2) To avoid preferring the few most discriminative pixels while neglecting other valuable pixel semantics in the specific mask, Gaussian suppression $\mathcal{F}_G$ is constructed to further adjust the attention distribution. Particularly, we select the top $K$ pixels with the highest scores as the discriminative pixels and construct the distance metric between these pixels and the surrounding neighboring pixels via the Gaussian kernel function. On this basis, a weighted aggregation of neighboring pixels replaces the original center pixel scores to suppress discriminative regions in the distribution, thus allowing the attention to expand to neighboring regions. Such a gentle manner is effective in enabling global semantic migration to aggregate richer class-specific semantics to prompts. At this time, the attention weight $\widetilde{A}_i\left (P,X\right)$ can be formulated as:
\begin{equation}\label{eq:5}
\setlength{\abovecaptionskip}{1pt}
\setlength{\belowcaptionskip}{1pt}
\widetilde{A}_{i(u,v)}=  \sum_{m,n\in \Upsilon } \mathcal{G}_{m,n}(\widetilde{A}_{i(u-m,v-n)}), \ s.t.\ \underset{u,v}{TopK} (\widetilde{A}_i)
\end{equation}
\renewcommand{\thefootnote}{\fnsymbol{footnote}}
\noindent where $(u,v)$ denotes the coordinates of top $K$ (i.e., set to 15) discriminative pixels, $\Upsilon \in [-5,5]$ denotes the neighborhood space and $\mathcal{G}_{m,n}$ denotes Gaussian kernel function\footnote[1]{Gaussian kernel function is a distance metric function, i.e., $\mathcal{G}_{m,n}=\frac{1}{2\pi\sigma^2 } \exp(-\frac{m^2+n^2}{2\sigma^2})$}. 

Therefore, based on the above improvements, SPT can transfer the part-level semantics within different part masks to prompts by:
\begin{equation}\label{eq:6}
\setlength{\abovecaptionskip}{1pt}
\setlength{\belowcaptionskip}{1pt}
\widetilde{P}=\left ( 1+ \mathcal{F}_{proj}\right ) \left ( \mathrm{softmax}\left ( \widetilde{A}\left (P,X\right) \right )\left ( XW_v \right ) \right )
\end{equation}
where $\mathcal{F}_{Proj}$ is formed by a two-layer MLP and $W_v$ denotes the projection weight. $\widetilde{P} = \left [ \widetilde{p}_1,\widetilde{p}_2,...,\widetilde{p}_{N_p} \right ] \in \mathbb{R}{^{N_p \times d}}$ denotes the enhanced class-aware FG-prompts, in which $\widetilde{p}_i$ aggregates the semantics from $i^{th}$ part mask region $m_i^p$. 

The BG-Prompt Enhancement follows a similar process, with the only difference being that the background masks (foreground masks reversed) are fed directly into the Semantic Prompt Transfer (SPT) to generate BG-prompts without Part Mask generator (PMG).

In summary, during the Prompt Enhancement, the prompts will aggregate class-specific semantics from the support/query image to generate corresponding class-aware prompts, which can be formulated as: 
\begin{equation}\label{eq:7}
\setlength{\abovecaptionskip}{1pt}
\setlength{\belowcaptionskip}{1pt}
\begin{split}
\widetilde{P}^q,\widetilde{B}^q& = \mathrm{PE}\left ( X^q,\hat{M}^q,P^q,B^q\right ) \\
\widetilde{P}^s,\widetilde{B}^s& = \mathrm{PE}\left (X^s,M^s,P^s,B^s\right )
\end{split}
\end{equation}
where $\mathrm{PE}$ denotes the entire Prompt Enhancement process via Eq.(\ref{eq:2})-Eq.(\ref{eq:6}). $X^q(X^s)$ denotes the query (support) feature. $\hat{M^q}$ denotes the pseudo query mask while $M_s$ denotes the support mask. $\widetilde{P}^*(\widetilde{B}^*)$ denotes the enhanced FG-prompts (BG-prompts). Notably, given that deep features are more class-specific, we choose to perform Prompt Enhancement in the last $L$ blocks of the encoder.

\subsection{Prompting} \label{Prompting}
After the Prompt Enhancement via Eq.(\ref{eq:7}), the enhanced FG-prompts and BG-prompts will interact with image features in the next encoder blocks. Given the similarity of image-level background information \cite{fu2023}, e.g., the "airplane" tends to appear in the air and the "ship" appears in the sea, we synchronize the support BG-prompts and query BG-prompts to generate generic BG-prompts. On the other hand, given the image-specific nature of the part-level semantics, FG-prompts interact with the corresponding image features independently without synchronization. The whole process of prompting can be formulated as:
\begin{equation}\label{eq:8}
\setlength{\abovecaptionskip}{1pt}
\setlength{\belowcaptionskip}{1pt}
\begin{split}
B_{n-1} &= \left ( B_{n-1}^q + B_{n-1}^s \right )/2 \\
\left [X_{n}^{q};{P_{n}^q};{B_{n}^q} \right ] &= E_n\left ( \left [X_{n-1}^q;P_{n-1}^q;B_{n-1}\right ]\right ) \\
\left [X_{n}^{s};{P_{n}^s};{B_{n}^s}\right ]& = E_n\left ( \left [X_{n-1}^s;P_{n-1}^s;B_{n-1}\right ]\right ) 
\end{split}
\end{equation}
where $E_n$ denotes the $n$-layer encoder block and $\left[;\right ]$ denotes the feature concatenate operation.  $\left [ {P_{n}^*}; {B_{n}^*}\right ]$ denote the support/query prompts after independent interaction with support/query tokens $X^*_{n-1}$, in $n$-layer block. 

Thus, we can alternate the above prompting and transferring processes (Eq.(\ref{eq:7})-(\ref{eq:8})) $L$ times (i.e., the number of the Prompt Enhancements) in the encoder blocks to continuously activate the specific objects in the image features and further refine the class-specific semantic descriptions of the prompts.

\subsection{Matching} \label{Matching}
Therefore, after $N$-layer encoder blocks, we directly utilize enhanced class-aware prompts $\left [ P^q; P^s; B \right ]$ to segment the activated query feature. Concretely, we perform cosine similarity computation of the class-aware query feature $X^q$ with fused FG-prompts, i.e. $\overline{P} = \lambda \cdot P^q+\left (1-\lambda   \right )\cdot P^s = \left [ \overline{p}_1,\overline{p}_2,...,\overline{p}_{N_p} \right ] \in \mathbb{R}{^{N_p \times d}}$ ($\lambda$ is learnable), and averaged BG-prompt $\overline{b}=\frac{1}{\left | N_b \right |}\sum_{j=1}^{N_b}b_j$, to achieve the final prediction $\mathcal{P}$, which can be formulated as:
\begin{equation}\label{eq:9}
\setlength{\abovecaptionskip}{1pt}
\setlength{\belowcaptionskip}{1pt}
{ \mathcal{P}_{(u,v)}}  =\left [ \left \langle X^q_{(u,v)},\overline{b}  \right \rangle  ;{ \max_{i \in \left [ 1,N_p \right ]  }} \left \{ \left \langle X^q_{(u,v)},\overline{p}_i \right \rangle     \right \}  \right ] 
\end{equation}
where $\left \langle \cdot,\cdot  \right \rangle $ denotes the cosine similarity. Benefiting from the part-level local semantics of FG-prompts, we employ the $\max$ operation to find the closest prompt to the feature pixel for part-level feature matching, which better handles the inter-image differences and region-level differences within images.

Thus, we utilize the cross-entropy loss $\mathcal{L}_{pred}$ to supervise the optimization of $\mathcal{P} $ towards the ground-truth label $M^q$. Furthermore, two losses $\mathcal{L}_{reg}$ and $\mathcal{L}_{dis}$ are proposed to capture more representative semantic prompts: 

1) We propose a loss of part-level regularization for preventing the part masks $M^p$ (include query $M^{q;p}$ and support $M^{s;p}$) from all gathering in the most salient target regions, thus generating similar part semantics: 
\begin{small}
\begin{equation}\label{eq:10}
\setlength{\abovecaptionskip}{1pt}
\setlength{\belowcaptionskip}{1pt}
{ \mathcal{L}_{reg}} =\frac{\sum_{i=1}^{N_p}\sum_{j\neq i\ }^{N_p} \left [ { \left \langle m^{q;p}_{i}, m^{q;p}_{j} \right \rangle  + \left \langle m^{s;p}_{i}, m^{s;p}_{j} \right \rangle }  \right ]  }{2*N_p*\left(N_p-1\right)}   
\end{equation}  
\end{small}

\noindent By minimizing the similarity between different part masks, the part masks are restricted to focus on different regions thus capturing different part prompts. 

2) We propose a prompt-based contrast loss to efficiently distinguish foreground semantics from background semantics: 
\begin{small}
\begin{equation}\label{eq:11}
\setlength{\abovecaptionskip}{1pt}
\setlength{\belowcaptionskip}{1pt}
{ \mathcal{L}_{dis}} = -{\log}  {\frac{\exp{\left \langle P^*,w_f^* \right \rangle}+\exp{ \left \langle B^*,w_b^* \right \rangle} }{\exp{\left \langle P^*,B^*  \right \rangle} +\exp{ \left \langle P^*,w_b^* \right \rangle} +\exp{ \left \langle B^*,w_f^*  \right \rangle } }  } 
\end{equation}  
\end{small}

\noindent where $w_f^*$ and $w_b^*$ are obtained by performing the masked average pooling operation from image features (query or support) with the ground-truth foreground mask and background mask, respectively. $P^*$ and $B^*$ are the FG-prompts and BG-prompts (query or support), respectively.

Overall, the training loss is:
\begin{equation}\label{eq:12}
\setlength{\abovecaptionskip}{1pt}
\setlength{\belowcaptionskip}{1pt}
\mathcal{L}=\mathcal{L}_{pred}+\alpha \cdot \mathcal{L}_{reg}+\beta \cdot \mathcal{L}_{dis}
\end{equation}  
where $\alpha$ and $\beta$ are loss weights to balance the corresponding contributions.

\begin{algorithm}
\setlength{\abovecaptionskip}{1pt}
\setlength{\belowcaptionskip}{1pt}
\caption{The overall process of our PAT}\label{algorithm}
\KwIn{ Support image with mask $\left (I^s, M^s, M^t \right )$ \\
\qquad \quad \, Query image $I^q$}
\KwOut{ Segmentation result $\mathcal{P}$}
Generate image tokens via the patch embedding layer $X^s \leftarrow I^s; X^q \leftarrow I^q$\;

\textbf{Prompt Initialization}: 
Derive initial FG-prompts and BG-prompts $\left (P,B \right) \leftarrow M^t $ according to Eq.(\ref{eq:1});

Feed the tokens and initial prompts to the Feature Encoder $E$ which contains $N$ blocks;

\For{$E_n$ in $E$}{
\If{N-n $\le$ L}{
\textbf{Prompt Enhancement}:
Transfer the class-specific semantics from image features to prompts according to Eq.(\ref{eq:7});
}
\textbf{Prompting}: The tokens interact with the prompts to activate the object according to Eq.(\ref{eq:8});
}

\textbf{Matching}: Calculate the similarity between the enhanced query feature with prompts $\mathcal{P} \leftarrow \left (X^q, P, B \right )$\ according to Eq.(\ref{eq:9}).
\end{algorithm}

\begin{table*}[t]
\setlength{\abovecaptionskip}{1pt}
\setlength{\belowcaptionskip}{1pt}
\caption{Performance on PASCAL-5$^i$ benchmark in terms of mIoU and FB-IoU ({\%}). ``-" indicates that the result is not recorded. ``Mean" and ``FB-IoU" indicate the average mIoU scores and average FB-IoU scores for four folds, respectively.} \label{tab:1}
\renewcommand\arraystretch{1.3}
\centering
\resizebox{0.92\linewidth}{!}{
\begin{tabular}{cr|cccc|cc|cccc|cc} 
\toprule
\multirow{2}{*}{Backbone}   & \multirow{2}{*}{Methond}                          & \multicolumn{6}{c|}{1-shot}                                                                                                                                                          & \multicolumn{6}{c}{5-shot}                                                                                                                \\ 
\cline{3-14}
&    & Fold-0      & Fold-1   & Fold-2  & Fold-3   & Mean & FB-IoU    & Fold-0                                             & Fold-1   & Fold-2    & Fold-3  & Mean    & FB-IoU                                              \\ 
\hline
\multirow{12}{*}{ResNet-50} & PFENet (TPAMI'2020)\cite{tian2020prior}                                            & 61.70                                              & 69.50                                              & 55.40                                              & 56.30                                              & 60.73                                              & 73.30                                              & 63.10                                              & 70.70                                              & 55.80                                              & 57.90                                              & 61.88                                              & 73.90                                               \\
& CyCTR (NeurIPS'2021)\cite{zhang2021few}                                             & 65.70                                              & 71.00                                              & 59.50                                              & 59.70                                              & 63.98                                              & -                                                  & 69.30                                              & 73.50                                              & 63.80                                              & 63.50                                              & 67.53                                              & -                                                   \\
& HSNet (ICCV'2020)\cite{min2021hypercorrelation}                                            & 64.30                                              & 70.70                                              & 60.30                                              & 60.50                                              & 63.95                                              & 76.70                                              & 70.30                                              & 73.20                                              & 67.40                                              & 67.10                                              & 69.50                                              & 80.60                                               \\
                
& HPA (TPAMI'2022)\cite{cheng2022holistic}                                               & 65.94                                              & 71.96                                              & 64.66                                              & 56.78                                              & 64.84                                              & 76.40                                              & 70.54                                              & 73.28                                              & 68.37                                              & 63.41                                              & 68.90                                              & 81.10                                               \\

& AAFormer (ECCV'2022)\cite{wang2022adaptive}                                          & 69.10                                              & 73.30                                              & 59.10                                              & 59.20                                              & 65.18                                              & 73.80                                              & 72.50                                              & 74.70                                              & 62.00                                              & 61.30                                              & 67.63                                              & 76.2                                                \\
& NERTNet (CVPR'2022)\cite{liu2022learning}                                           & 65.40                                              & 72.30                                              & 59.40                                              & 59.80                                              & 64.23                                              & 77.00                                              & 66.20                                              & 72.80                                              & 61.70                                              & 62.20                                              & 65.73                                              & 78.40                                               \\
& BAM (CVPR'2022)\cite{lang2022learning}                                               & 68.97                                              & 73.59                                              & 67.55                                              & 61.13                                              & 67.81                                              & 79.71                                              & 70.59                                              & 75.05                                              & 70.79                                              & 67.20                                              & 70.91                                              & 82.20                                               \\
& SVF (NeurIPS'2022)\cite{sun2022singular}
& 69.38 &74.51 & 68.80 & 63.09 &68.95 &- &72.05 &76.17 & 71.97 &68.91 & 72.28 &- \\ 
& IPMT (NeurIPS'2022)\cite{liu2022intermediate}                                              & 72.80                                              & 73.70                                              & 59.20                                              & 61.60                                              & 66.83                                              & 77.10                                              & 73.10                                              & 74.70                                              & 61.60                                              & 63.40                                              & 68.20                                              & 81.40                                               \\

& ASNet (CVPR'2023)\cite{kang2022integrative}                                           & 68.90                                              & 71.70                                              & 61.10                                              & 62.70                                              & 66.10                                              & 77.70                                              & 72.60                                              & 74.30                                              & 65.30                                              & 67.10                                              & 69.83                                              & 80.40                                               \\
& HDMNet (CVPR'2023)\cite{peng2023hierarchical}                                            & 71.00                                              & 75.40                                              & 68.90                                              & 62.10                                              & 69.35                                              & -                                                  & 71.30                                              & 76.20                                              & 71.30                                              & 68.50                                              & 71.83                                              & -                                                   \\
& MIANet (CVPR'2023)\cite{yang2023mianet}                                            & 68.51                                              & 75.76                                              & 67.46                                              & 63.15                                              & 68.72                                              & -                                                  & 70.20                                              & 77.38                                              & 70.02                                              & 68.77                                              & 71.59                                              & -                                                   \\
& RiFeNet (AAAI'2024)\cite{bao2023relevant}                                            & 68.40                                             & 73.50                                             & 67.10                                              & 59.40                                              & 67.10                                              & -                                                  & 70.00                                              & 74.70                                             & 69.40                                              & 64.20                                              & 69.60                                              & -                                                   \\
& PMNet (WACV'2024)\cite{chen2024pixel}                                            & 67.30                                             & 72.00                                             & 62.40                                              & 59.90                                              & 65.40                                              & -                                                  & 73.60                                              & 74.60                                             & 69.90                                              & 67.20                                              & 71.30                                              & -                                                   \\

\hline
\multirow{9}{*}{ResNet-101} 

& CyCTR (NeurIPS'2021)\cite{zhang2021few}                                            & 67.20                                              & 71.10                                              & 57.60                                              & 59.00                                              & 63.73                                              & 72.90                                              & 71.00                                              & 75.00                                              & 58.50                                              & 65.00                                              & 67.38                                              & 75.00                                               \\
& HSNet (ICCV'2021)\cite{min2021hypercorrelation}                                            & 67.30                                              & 72.30                                              & 62.00                                              & 63.10                                              & 66.18                                              & 77.60                                              & 71.80                                              & 74.40                                              & 67.00                                              & 68.30                                              & 70.38                                              & 80.60                                               \\
& APANet (TMM'2022)\cite{chen2021apanet}                                              & 63.10                                              & 71.10                                              & 63.80                                              & 57.90                                              & 63.98                                              & 74.90                                              & 67.50                                              & 73.30                                              & 67.90                                              & 63.10                                              & 67.95                                              & 78.80                                               \\                            
& HPA (TPAMI'2022)\cite{cheng2022holistic}                                              & 66.40                                              & 72.65                                              & 64.10                                              & 59.42                                              & 65.64                                              & 76.60                                              & 68.02                                              & 74.60                                              & 65.85                                              & 67.11                                              & 68.90                                              & 80.40                                               \\

& SSPNet (ECCV'2022)\cite{fan2022self}                                            & 63.70                                              & 70.10                                              & 66.70                                              & 55.40                                              & 63.98                                              & -                                                  & 70.30                                              & 76.30                                              & 77.80                                              & 65.50                                              & 72.48                                              & -                                                   \\
& AAFormer (ECCV'2022)\cite{wang2022adaptive}                                          & 69.90                                              & 73.60                                              & 57.90                                              & 59.70                                              & 65.28                                              & 74.90                                              & 75.00                                              & 75.10                                              & 59.00                                              & 63.20                                              & 68.08                                              & 77.30                                               \\
& NERTNet (CVPR'2022)\cite{liu2022learning}                                            & 65.50                                              & 71.80                                              & 59.10                                              & 58.30                                              & 63.68                                              & 75.30                                              & 67.90                                              & 73.20                                              & 60.10                                              & 66.80                                              & 67.00                                              & 78.20                                               \\
& IPMT (NeurIPS'2022)\cite{liu2022intermediate}                                             & 71.60                                              & 73.50                                              & 58.00                                              & 61.20                                              & 66.08                                              & 78.50                                              & 75.30                                              & 76.90                                              & 59.60                                              & 65.10                                              & 69.23                                              & 80.30                                               \\ 
& RiFeNet (AAAI'2024)\cite{bao2023relevant}                                            & 68.90                                             & 73.80                                             & 66.20                                              & 60.30                                              & 67.30                                              & -                                                  & 70.40                                              & 74.50                                             & 68.30                                              & 63.40                                              & 69.20                                              & -                                                   \\
& PMNet (WACV'2024)\cite{chen2024pixel}                                            & 71.30                                             & 72.40                                             & 66.90                                              & 61.90                                              & 68.10                                              & -                                                  & 74.90                                              & 75.50                                             & 75.30                                              & 69.80                                              & 73.90                                              & -                                                   \\
\hline
\multirow{2}{*}{ViT-B/16}   & Fptrans (NeurIPS'2022)\cite{zhang2022feature}                                          & 67.10                                              & 69.80                                              & 65.60                                              & 56.40                                              & 64.73                                              & -                                                  & 73.50                                              & 75.70                                              & 77.40                                              & 68.30                                              & 73.73                                              & -                                                   \\
& {\cellcolor{gray!15}}\textbf{PAT (Ours)} & {\cellcolor{gray!15}}\textbf{68.33} & {\cellcolor{gray!15}}\textbf{73.24} & {\cellcolor{gray!15}}\textbf{66.16} & {\cellcolor{gray!15}}\textbf{60.11} & {\cellcolor{gray!15}}\textbf{66.96} & {\cellcolor{gray!15}}\textbf{78.81} & {\cellcolor{gray!15}}\textbf{73.33} & {\cellcolor{gray!15}}\textbf{77.61} & {\cellcolor{gray!15}}\textbf{75.12} & {\cellcolor{gray!15}}\textbf{69.45}          & {\cellcolor{gray!15}}\textbf{73.88}          & {\cellcolor{gray!15}}\textbf{83.55}  \\ 
\hline
\multirow{3}{*}{DeiT-B/16}  & Fptrans (NeurIPS'2022)\cite{zhang2022feature}                                          & 72.30                                              & 70.60                                              & 68.30                                              & 64.10                                              & 68.83                                              & -                                                  & 76.70                                              & 79.00                                              & 81.00                                              & 75.10                                              & 77.95                                              & -                                                   \\
& MuHS (ICLR'2023)\cite{hu2022suppressing}                                             & 71.20                                              & 71.50                                              & 67.00                                              & 66.60                                              & 69.08                                              & -                                                  & 75.70                                              & 77.80                                              & 78.60                                              & 74.70                                              & 76.70                                              & -                                                   \\
& {\cellcolor{gray!15}}\textbf{PAT (Ours)} & {\cellcolor{gray!15}}\textbf{71.71} & {\cellcolor{gray!15}}\textbf{76.85} & {\cellcolor{gray!15}}\textbf{67.56} & {\cellcolor{gray!15}}\textbf{70.51} & {\cellcolor{gray!15}}\textbf{71.66} & {\cellcolor{gray!15}}\textbf{81.59} & {\cellcolor{gray!15}}\textbf{76.90} & {\cellcolor{gray!15}}\textbf{80.26} & {\cellcolor{gray!15}}\textbf{78.13} & {\cellcolor{gray!15}}\textbf{76.76} & {\cellcolor{gray!15}}\textbf{78.01} & {\cellcolor[rgb]{0.922,0.925,0.925}}\textbf{86.43}  \\
\bottomrule
\end{tabular}}
\end{table*}

\begin{figure*}[t]
\centering
\setlength{\abovecaptionskip}{1pt}
\setlength{\belowcaptionskip}{1pt}
\includegraphics[width=0.9\textwidth]{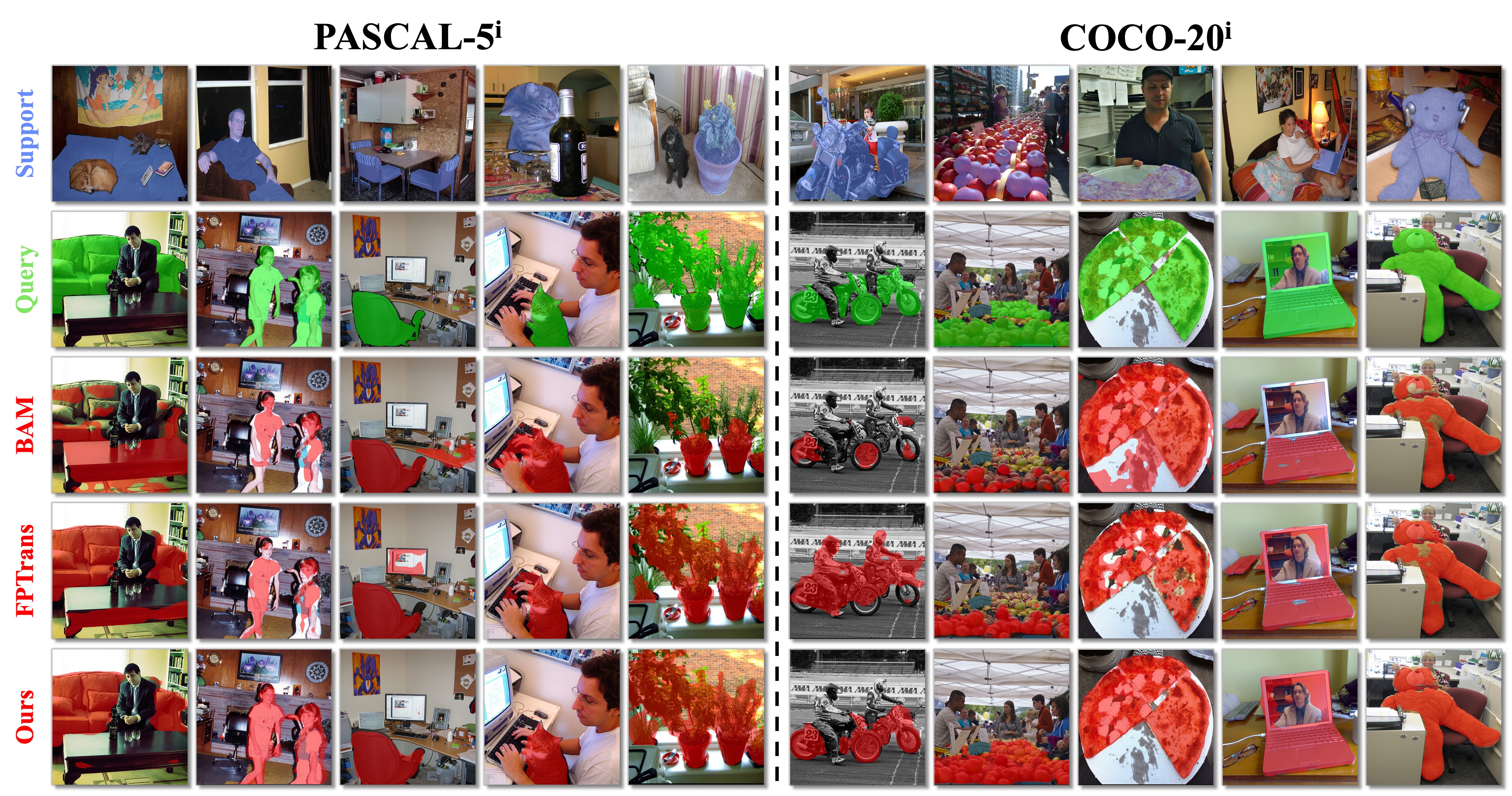} 
\caption{Qualitative results of the proposed method and several competitive methods under the 1-shot setting. Results in the left panel are from PASCAL-5$^i$ while those in the right panel are from COCO-20$^i$. Each row from the top to the bottom represents: support images with GT (\textcolor{blue}{blue}), query images with GT (\textcolor{green}{green}), results of BAM (\textcolor{red}{red}), results of FPTrans (\textcolor{red}{red}), and our results (\textcolor{red}{red}).}
\label{fig:6}
\end{figure*}

\begin{figure}[t]
\centering
\setlength{\abovecaptionskip}{1pt}
\setlength{\belowcaptionskip}{1pt}
\includegraphics[width=0.95\linewidth]{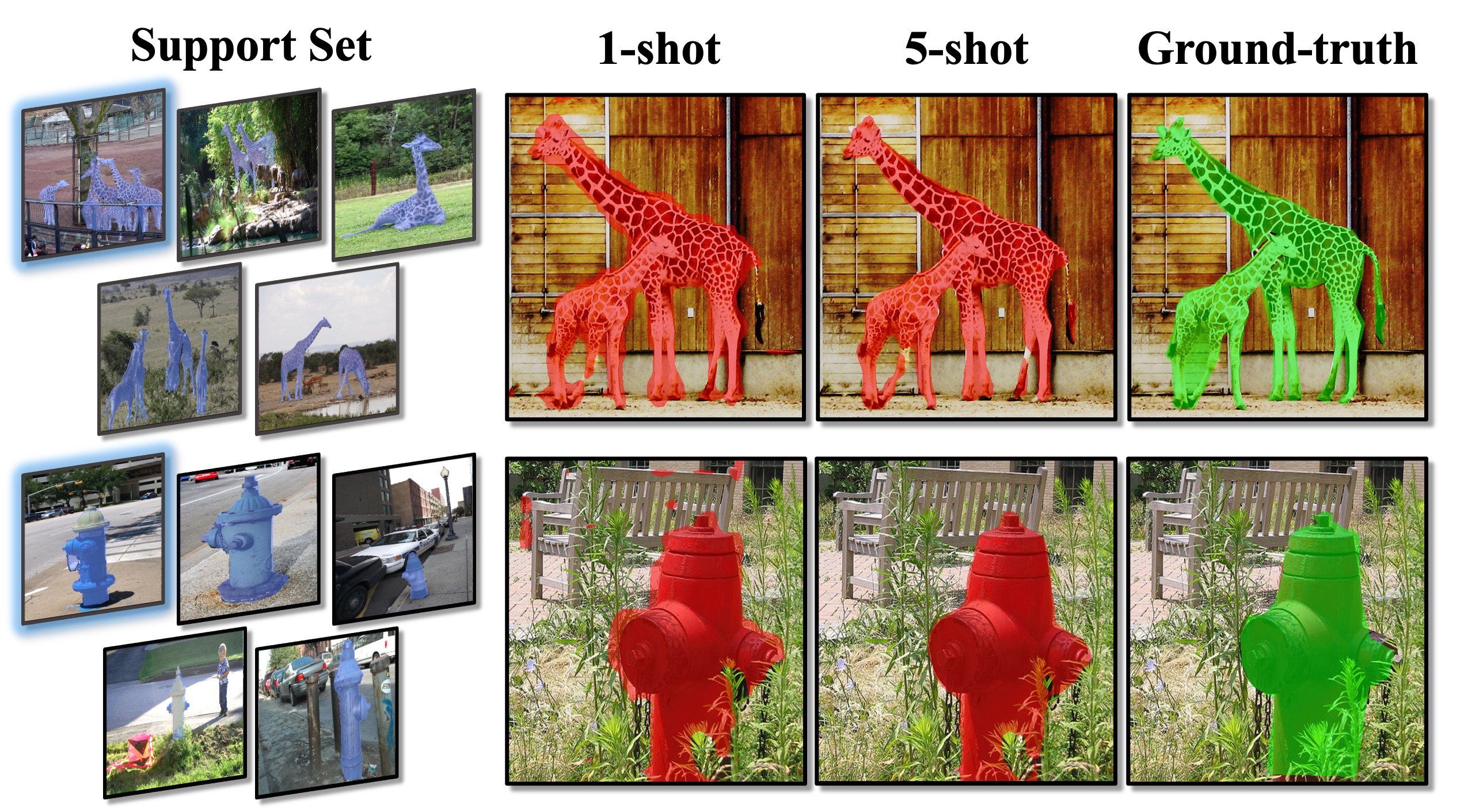}
\caption{Comparison results of our method under the 1- and 5-shot settings on COCO-20$^i$ dataset. The first labeled image (blue border) in the Support Set is sampled for 1-shot segmentation.} \label{fig:7}
\end{figure} 

\begin{figure}[t]
\centering
\setlength{\abovecaptionskip}{1pt}
\setlength{\belowcaptionskip}{1pt}
\includegraphics[width=0.95\linewidth]{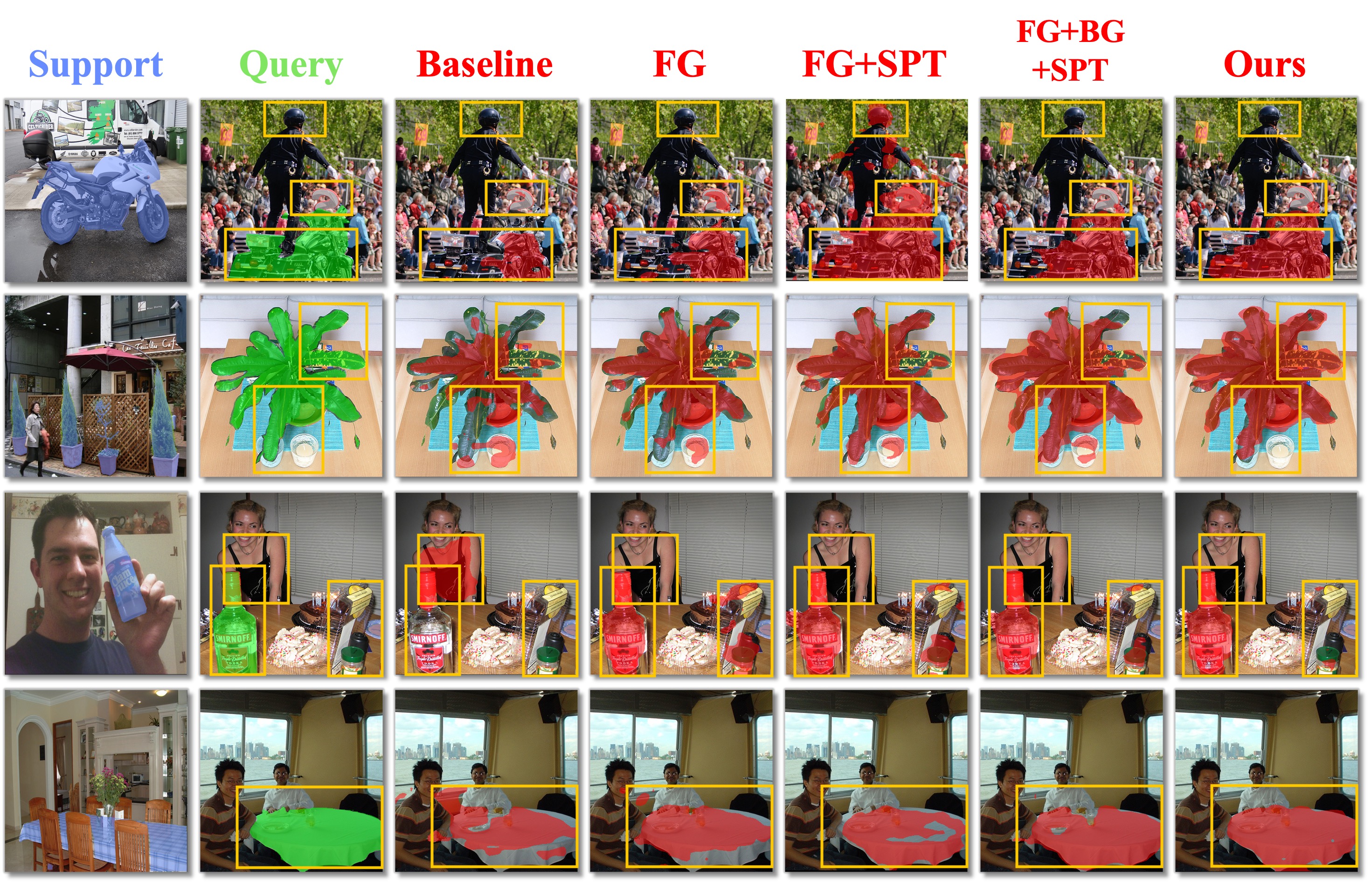}
\caption{Qualitative Comparison of the variants of PAT. Each column from left to right: support images (GT), query images (GT), results of baseline, results of exploiting FG-prompts, results of exploiting FG-prompts+SPT, results of exploiting FG-prompts+BG-prompts+SPT, results of exploiting FG-prompts+BG-prompts+SPT+PMG (i.e., PAT).} 
\label{fig:8}
\end{figure}

\section{Experiments}
In this section, we demonstrate the performance of our method on four different tasks. Firstly, we describe the experimental setup for the standard FSS in Sec.\ref{Experimental Setup}. Then Sec.\ref{Comparison with the State of-the-Arts} and Sec.\ref{Ablation Studies} describe the performance comparisons with other methods and ablation studies, respectively, on the standard FSS task. Sec.\ref{Extended Experiments} demonstrates the performance of the proposed PAT on several more challenging extended tasks. Finally, some failure cases are described in Sec.\ref{Failure case analysis}.

\subsection{Experimental Setup} \label{Experimental Setup}
\subsubsection{Datasets and Evaluation Metric} 
We evaluate the proposed method on three popular FSS datasets PASCAL-5$^i$ \cite{shaban2017one}, COCO-20$^i$ \cite{nguyen2019feature} and iSAID \cite{bi2023not}. PASCAL-5$^i$ is composed of PASCAL VOC 2012 \cite{everingham2010pascal} and SBD \cite{hariharan2011semantic}, which includes 20 classes. COCO-20$^i$ is constructed from MSCOCO \cite{lin2014microsoft}, which includes 80 classes. We divide the two datasets into 4 folds, each fold in PASCAL-5$^i$ contains 15 classes for training and 5 classes for testing, while in COCO-20$^i$ contains 60 classes for training and 20 classes for testing. For each fold, we randomly sample 5000 image pairs for validation. In addition, we perform the FSS task on a remote sensing dataset, i.e., iSAID~\cite{waqas2019isaid}, which includes 15 classes. We divide the dataset into 3 folds, each containing 10 classes for training and 5 classes for testing.

For a fair comparison, mean intersection-over-union (mIoU) and foreground-background IoU (FB-IoU) are employed as evaluation metrics \cite{lang2022learning,tian2020prior}.

\subsubsection{Implementation details}
The proposed model is implemented on Tesla A40 GPUs by using the PyTorch framework. We employ two vision transformers, DeiT-B/16 \cite{touvron2021training} and ViT-B/16 \cite{dosovitskiy2020image}, as encoders due to their unique attention structure to flexibly insert inputs (prompts). Both of them are pre-trained on ImageNet \cite{russakovsky2015imagenet}. The batch size is set to 4 and the SGD optimizer is employed to update the parameters of the model. Following the previous FSS setups \cite{lang2022learning, bi2023not}, we randomly crop the images to 480$\times$480 (iSAID cropped to 256$\times$256) and adopt the same data augmentation strategies. For PASCAL-5$^i$, we train the model for 50 epochs and set the initial learning rate to 1e-3. For COCO-20$^i$, we train the model for 40 epochs and set the initial learning rate to 6e-4. For iSAID, we train the model for 40 epochs and set the initial learning rate to 1e-3. We set both the number of FG-prompts $N_p$ and BG-prompts $N_b$ to 8. The factors $\alpha$ and $\beta$ for balancing the loss contribution in Eq.(\ref{eq:12}) are set to 5e-2. 

\subsection{Comparison with the State of-the-Arts} \label{Comparison with the State of-the-Arts}
\subsubsection{Quantitative Analysis}
\textbf{PASCAL-5$^i$: } 
Table \ref{tab:1} demonstrates the performance comparison of different FSS methods on PASCAL-5$^i$ benchmark, where our PAT achieves state-of-the-art performance in all cases. With the backbone of DeiT-B/16, PAT surpasses the most competitive CNN-based HDMNet \cite{peng2023hierarchical} by 2.31{\%} (1-shot) and 6.18{\%} (5-shot) mIoU, respectively. Notably, our 1-shot result almost achieves the 5-shot result of HDMNet, i.e., 71.66{\%} $vs.$ 71.83{\%}, indicating its superiority in effectively minimizing the number of support samples required for FSS. In addition, the proposed PAT also achieves a 2.58{\%} mIoU increment compared to two transformer-based methods, FPTrans \cite{zhang2022feature}, and MuHS \cite{hu2022suppressing}, under the 1-shot setting. Moreover, the 1.88{\%} (1-shot) and 4.23{\%} (5-shot) FB-IoU increments similarly demonstrate its superiority.

\noindent \textbf{COCO-20$^i$: } Table \ref{tab:2} presents the performance on COCO-20$^i$ benchmark. We can find that our PAT also achieves competitive segmentation performance when faced with more complex and diverse scenarios. In particular, our method achieves the 48.05{\%} mIoU score, outperforming the best CNN-based competitor and the best transformer-based competitor by 1.82{\%} and 0.7{\%} mIoU, respectively. Furthermore, in terms of FB-IoU, our PAT beats the state-of-the-arts by 0.34{\%} and 2.38{\%} FB-IoU under the 1- and 5-shot settings, respectively.

\noindent \textbf{iSAID: } Table \ref{tab:4_0} demonstrates the performance on iSAID benchmark. Our PAT achieves the best segmentation performance even in remote sensing scenarios with larger intra- and inter-class variance, with 50.51\% and 59.41\% mIoU under the 1- and 5-shot settings, respectively.

\subsubsection{Qualitative Analysis}
To further demonstrate the superiority of the proposed method, we visualize the corresponding segmentation results in Fig.\ref{fig:6}. Our PAT exhibits several advantages over other competitors: (1) PAT can efficiently focus on the novel class in a dynamic class-aware prompting manner, and significantly suppress false activation of base classes (e.g., the "Chair" in $1^{st}$ column, the "Person" in the $6^{th}$ column). (2) By mining the part-level prompts, PAT can construct fine-grained part-level feature matching to tackle region-level differences within images and inter-image differences, while other competitors are prone to produce imprecise and incomplete results (e.g., "Potted plant" in $5^{th}$ column and "Apple" in $7^{th}$ column). In addition, Fig.\ref{fig:7} compares the performance of our PAT under the 1- and 5-shot settings. It can be found that the 5-shot results are superior to the 1-shot results with more accurate boundaries and complete objects. The insight behind the phenomenon is that the class-aware prompts derived from aggregating multiple support image clues are more discriminative and representative.

\begin{table*}[t]
\setlength{\abovecaptionskip}{1pt}
\setlength{\belowcaptionskip}{1pt}
\caption{ Performance on COCO-20$^i$ benchmark in terms of mIoU and FB-IoU ({\%}). ``-" indicates that the result is not recorded. ``Mean" and ``FB-IoU" indicate the average mIoU scores and average FB-IoU scores for four folds, respectively.} \label{tab:2}
\renewcommand\arraystretch{1.3}
\centering
\resizebox{0.9\linewidth}{!}{
\begin{tabular}{cr|cccc|cc|cccc|cc} 
\toprule
\multirow{2}{*}{Backbone}   & \multirow{2}{*}{Methond} & \multicolumn{6}{c|}{1-shot}                                                                          & \multicolumn{6}{c}{5-shot}                                                                            \\ 
\cline{3-14}
&                          & Fold-0         & Fold-1         & Fold-2         & Fold-3         & Mean           & FB-IoU         & Fold-0         & Fold-1         & Fold-2         & Fold-3         & Mean           & FB-IoU          \\ 
\hline
\multirow{11}{*}{ResNet-50} 

& CyCTR (NeurIPS'2021)\cite{zhang2021few}         & 38.90          & 43.00          & 39.60          & 39.80          & 40.33          & -              & 41.10          & 48.90          & 45.20          & 47.00          & 45.55          & -               \\
& HSNet (ICCV'2021)\cite{min2021hypercorrelation}           & 36.30          & 43.10          & 38.70          & 38.70          & 39.20          & 68.20          & 43.30          & 51.30          & 48.20          & 45.00          & 46.95          & 70.70           \\
& HPA (TPAMI'2022)\cite{cheng2022holistic}            & 40.30          & 46.57          & 44.12          & 42.71          & 43.43          & 68.21          & 45.54          & 55.43          & 48.90          & 50.21          & 50.02          & 71.15           \\                     
& AAFormer (ECCV'2022)\cite{wang2022adaptive}    & 39.80          & 44.60          & 40.60          & 41.40          & 41.60          & 67.70          & 42.90          & 50.10          & 45.50          & 49.20          & 46.93          & 68.20           \\
& NERTNet (CVPR'2022)\cite{liu2022learning}         & 36.80          & 42.60          & 39.90          & 37.90          & 39.30          & 68.50          & 38.20          & 44.10          & 40.40          & 38.40          & 40.28          & 69.20           \\
& BAM (CVPR'2022)\cite{lang2022learning}              & 43.41          & 50.59          & 47.49          & 43.43          & 46.23          & -              & 49.26          & 54.20          & 51.63          & 49.55          & 51.16          & -               \\
& IPMT (NeurIPS'2022)\cite{liu2022intermediate}         & 41.40          & 45.10          & 45.60          & 40.00          & 43.03          & -              & 43.50          & 49.70          & 48.70          & 47.90          & 47.45          & -               \\
& ASNet (CVPR'2023)\cite{kang2022integrative}            & -              & -              & -              & -              & 42.20          & 68.80          & -              & -              & -              & -              & 47.90          & 71.60           \\
& ABCNet 
(CVPR'2023)\cite{wang2023rethinking}           & 40.70              & 45.90              & 41.60              & 40.60             & 42.20          & 66.70          & 43.20              & 50.80              & 45.80              & 47.10              & 46.70          & 62.80           \\      
& RiFeNet 
(AAAI'2024)\cite{bao2023relevant}           & 39.10              & 47.20              & 44.60              & 45.40             & 44.10          & -          & 44.30              & 52.40              & 49.30              & 48.40              & 48.60          & -           \\                               

& PMNet (WACV'2024)\cite{chen2024pixel}                                            & 39.80                                            & 41.00                                             & 40.10                                              & 40.70                                              & 40.40                                              & -                                                  & 50.10                                              & 51.00                                             & 50.40                                              & 49.60                                              & 50.30                                              & -                                                   \\                            
\hline
\multirow{7}{*}{ResNet-101} 
& PFENet (TPAMI'2020)\cite{tian2020prior}         & 34.40          & 33.00          & 32.30          & 30.10          & 32.45          & 63.00          & 38.50          & 38.60          & 38.20          & 34.30          & 37.40          & 65.80           \\
& APANet (TMM'2022)\cite{chen2021apanet}              & 40.70          & 44.60          & 42.50          & 39.60          & 41.85          & 64.80          & 45.70          & 49.70          & 47.40          & 42.80          & 46.40          & 69.80           \\
& HPA (TPAMI'2022)\cite{cheng2022holistic}            & 43.08          & 50.01          & 44.78          & 45.19          & 45.77          & 68.37          & 49.18          & 57.76          & 51.97          & 50.64          & 52.39          & 74.02           \\                             
& SSPNet (ECCV'2022)\cite{fan2022self}          & 39.10          & 45.10          & 42.70          & 41.20          & 42.03          & -              & 47.40          & 54.50          & 50.40          & 49.60          & 50.48          & -               \\
& NTRENet (CVPR'2022)\cite{liu2022learning}         & 38.30          & 40.40          & 39.50          & 38.10          & 39.08          & 67.50          & 42.30          & 44.40          & 44.20          & 41.70          & 43.15          & 69.60           \\
& IPMT (NeurIPS'2022)\cite{liu2022intermediate}         & 40.50          & 45.70          & 44.80          & 39.30          & 42.58          & -              & 45.10          & 50.30          & 49.30          & 46.80          & 47.88          & -               \\
& PMNet (WACV'2024)\cite{chen2024pixel}                                            & 44.70                                             & 44.30                                             & 44.00                                              & 41.80                                              & 43.70                                              & -                                                  & 52.60                                              & 53.30                                             & 53.50                                              & 52.80                                              & 53.10                                              & -                                                   \\

\hline

\multirow{3}{*}{DeiT-B/16}  & Fptrans (NeurIPS'2022)\cite{zhang2022feature}      & 44.40          & 48.90          & 50.60          & 44.00          & 46.98          & -              & 54.20          & 62.50          & 61.30          & 57.60          & 58.90          & -               \\
& MuHS (ICLR'2023)\cite{hu2022suppressing}            & 44.00          & 50.00          & 49.10          & 46.30          & 47.35          & -              & 53.60          & 60.50          & 57.30          & 55.20          & 56.65          & -               \\

& {\cellcolor{gray!15}}\textbf{PAT (Ours)}& {\cellcolor{gray!15}}\textbf{40.62} & {\cellcolor{gray!15}}\textbf{51.94} & {\cellcolor{gray!15}}\textbf{48.99} & {\cellcolor{gray!15}}\textbf{50.66} & {\cellcolor{gray!15}}\textbf{48.05} & {\cellcolor{gray!15}}\textbf{69.14} & {\cellcolor{gray!15}}\textbf{51.26} & {\cellcolor{gray!15}}\textbf{63.15} & {\cellcolor{gray!15}}\textbf{59.45} & {\cellcolor{gray!15}}\textbf{55.88} & {\cellcolor{gray!15}}\textbf{57.44} & {\cellcolor{gray!15}}\textbf{76.40}  \\
\bottomrule
\end{tabular}}
\end{table*}

\begin{table*}[t]
\setlength{\abovecaptionskip}{1pt}
\setlength{\belowcaptionskip}{1pt}
\caption{Performance on iSAID \cite{waqas2019isaid} benchmark in terms of mIoU and FB-IoU ({\%}). "Mean" and "FB-IoU" indicate the average mIoU scores and average FB-IoU scores for three folds, respectively.} \label{tab:4_0}
\renewcommand\arraystretch{1.3}
\centering
\resizebox{0.85\linewidth}{!}{
\begin{tabular}{cr|ccc|cc|ccc|cc} 
\toprule
\multirow{2}{*}{Backbone}    & \multirow{2}{*}{Method}                   & \multicolumn{5}{c|}{1-shot}   & \multicolumn{5}{c}{5-shot}      \\ 
\cline{3-12}
&                                                  & Fold-0                                             & Fold-1                                             & Fold-2                                             & Mean                                              & FB-IoU         & Fold-0                                             & Fold-1                                             & Fold-2                                             & Mean                                             & FB-IoU          \\ 
\hline
\multirow{8}{*}{ResNet-50}  
& PFENet (TPAMI'2020)~\cite{tian2020prior}                              & 51.34                                              & 38.79                                              & 52.26                                              & 47.46                                              & 63.34          & 54.71                                              & 41.51                                              & 54.45                                              & 50.22                                              & 64.99           \\
& CyCTR (NeurIPS'2021)~\cite{zhang2021few}                            & 51.15                                              & 38.40                                              & 53.79                                              & 47.78                                              & 62.86          & 51.91                                              & 39.01                                              & 54.83                                              & 48.58                                              & 63.81           \\

& IPMT (NeurIPS'2021)~\cite{liu2022intermediate} & 53.96                & 38.76                               & 53.73          & 48.82                               & 63.68          & 54.50                & 43.40                               &56.71          & 51.54           & 64.69              \\
& NTRENet (CVPR'2022)~\cite{liu2022learning}                              & 49.52                                              & 38.66                                              & 51.87                                              & 46.68                                              & 62.84          & 50.60                                              & 40.99                                              & 55.07                                              & 48.89                                              & 63.74           \\
& DCPNet (IJCV'2023)~\cite{lang2024few}                               & 48.43                                              & 37.59                                              & 52.09                                              & 46.04                                              & 62.76          & 50.34                                              & 40.75                                              & 51.33                                              & 47.47                                              & 64.10           \\
& SCCAN (ICCV'2023)~\cite{xu2023self}                               & 52.24                                              & 38.60                                              & 52.84                                              & 47.89                                              & 63.26          & 52.75                                              & 38.95                                              & 54.26                                              & 48.65                                              & 62.23           \\
& R2Net (TGRS'2023)~\cite{lang2023global}   & 56.81  & 39.85 &49.02  &48.56 &62.65              &60.47                                                  &41.43  &50.24                                                  &50.71                                                  &65.70               \\
& DMNet (TGRS'2023)~\cite{bi2023not}                               & 54.45                                              & 40.68                                              & 53.60                                              & 49.58                                              & 64.46          & 57.67                                              & 41.06                                              & 55.28                                              & 51.34                                              & 65.81           \\ 
\hline
\multirow{8}{*}{ResNet-101} 

& PFENet (TPAMI'2020)~\cite{tian2020prior}                             & 50.69                                              & 38.37                                              & 52.85                                              & 47.30                                              & 62.46          & 54.40                                              & 41.55                                              & 50.55                                              & 48.83                                              & 64.57           \\
& CyCTR (NeurIPS'2021)~\cite{zhang2021few}                            & 50.89                                              & 38.89                                              & 52.22                                              & 47.33                                              & 62.35          & 52.15                                              & 40.28                                              & 55.32                                              & 49.25                                              & 64.45           \\
& IPMT (NeurIPS'2021)~\cite{liu2022intermediate} & 52.62                & 39.83                               & 54.10                              & 48.85                               & 63.19                              & 53.38                & 44.40                               & 54.50                              & 50.76           & 64.35              \\
& NTRENet (CVPR'2022)~\cite{liu2022learning}                             & 50.33                                              & 38.73                                              & 51.23                                              & 46.76                                              & 63.25          & 53.24                                              & 41.87                                              & 51.53                                              & 48.88                                              & 64.16           \\
& DCPNet (IJCV'2023)~\cite{lang2024few}                              & 47.63                                              & 38.80                                              & 49.34                                              & 45.26                                              & 62.56          & 50.68                                              & 40.02                                              & 54.52                                              & 48.41                                              & 62.96           \\
& SCCAN (ICCV'2023)~\cite{xu2023self}                               & 52.71                                              & 38.76                                              & 53.45                                              & 48.31                                              & 62.72          & 53.85                                              & 39.11                                              & 54.56                                              & 49.17                                              & 64.27           \\
& R2Net (TGRS'2023)~\cite{lang2023global}                               & 57.74  & 39.76 &48.45  &48.65 &63.38              &58.83                                                  &40.58  &49.69                                                  &49.70                                                  &63.79           \\
& DMNet (TGRS'2023)~\cite{bi2023not}                               & 54.01                                              & 40.04                                              & 53.57                                              & 49.21                                              & 64.03          & 55.70                                              & 41.69                                              & 56.47                                              & 51.29                                              & 65.88           \\ 
\hline
\multirow{2}{*}{DeiT-B/16}   & FPTrans (NeurIPS'2022)~\cite{zhang2022feature}                          & 51.59                                              & 37.79                                              & 52.85                                              & 47.41                                              & 62.52          & 58.52                                              & 47.24                                              & 65.37                                              & 57.04                                              & 67.63           \\
& {\cellcolor[rgb]{0.925,0.925,0.925}}\textbf{PAT (Ours)} & {\cellcolor[rgb]{0.925,0.925,0.925}}\textbf{54.81} & {\cellcolor[rgb]{0.925,0.925,0.925}}\textbf{43.03} & {\cellcolor[rgb]{0.925,0.925,0.925}}\textbf{53.70} & {\cellcolor[rgb]{0.925,0.925,0.925}}\textbf{50.51} & {\cellcolor[rgb]{0.925,0.925,0.925}}\textbf{64.47} & {\cellcolor[rgb]{0.925,0.925,0.925}}\textbf{62.40}  & {\cellcolor[rgb]{0.925,0.925,0.925}}\textbf{49.72} & {\cellcolor[rgb]{0.925,0.925,0.925}}\textbf{66.10}  & {\cellcolor[rgb]{0.925,0.925,0.925}}\textbf{59.41} & {\cellcolor[rgb]{0.925,0.925,0.925}}\textbf{70.20}  \\
\bottomrule
\end{tabular}}
\end{table*}

\begin{table}[t]
\setlength{\abovecaptionskip}{1pt}
\setlength{\belowcaptionskip}{1pt}
\caption{ Ablation studies of different components of our PAT under the 1-shot setting. ``FG" and ``BG" indicate FG-prompts and BG-prompts, respectively. ``SPT" indicates the Semantic Prompt Transfer and ``PMG" indicates the Part Mask Generator. \textbf{Bold} and \underline{underline} indicate the best and the sub-best performance, respectively.} \label{tab:8}
\renewcommand\arraystretch{1.3}
\centering
\resizebox{0.95\linewidth}{!}{
\begin{tabular}{c|cc|cc|cc|cc} 
\toprule
\multirow{2}{*}{\#} & \multicolumn{2}{c|}{Prompting} & \multicolumn{2}{c|}{Transferring}                     & \multicolumn{2}{c|}{PASCAL-$5^i$}     & \multicolumn{2}{c}{COCO-$20^i$}         \\ 
\cline{2-9}
& FG & BG                        & \multicolumn{1}{l}{SPT} & \multicolumn{1}{l|}{PMG} & mIoU           & FB-IoU         & mIoU           & FB-IoU        \\ 
\hline
(a)                 & \ding{55}   & \ding{55}                              & \ding{55}    & \ding{55}                   & 67.30 & 77.56               & 42.14 & 65.00             \\
(b)                 & \ding{51}  &  \ding{55}                             & \ding{55}    & \ding{55}                   & 68.26 & 78.81               & 43.75 & 65.16             \\
(c)                 & \ding{51}  & \ding{55}                              & \ding{51}   & \ding{55}                   & 69.57 & 79.58               & 45.58 & 67.82             \\
(d)                 & \ding{51}  & \ding{51}                             & \ding{51}   & \ding{55}                   & 70.52 & 80.40                & 46.71 & 67.87             \\
(e)                 & \ding{51}  & \ding{55}                              & \ding{51}   & \ding{51}                  & \underline{70.68} & \underline{80.96}               & \underline{46.92} & \underline{68.11}             \\
(f)                 & \ding{51}  & \ding{51}                             & \ding{51}   & \ding{51}                  & \textbf{71.66} & \textbf{81.59} & \textbf{48.05} & \textbf{69.14}             \\
\bottomrule
\end{tabular}}
\end{table}

\begin{figure}[t]
\centering
\setlength{\abovecaptionskip}{1pt}
\setlength{\belowcaptionskip}{1pt}
\includegraphics[width=1.0\linewidth]{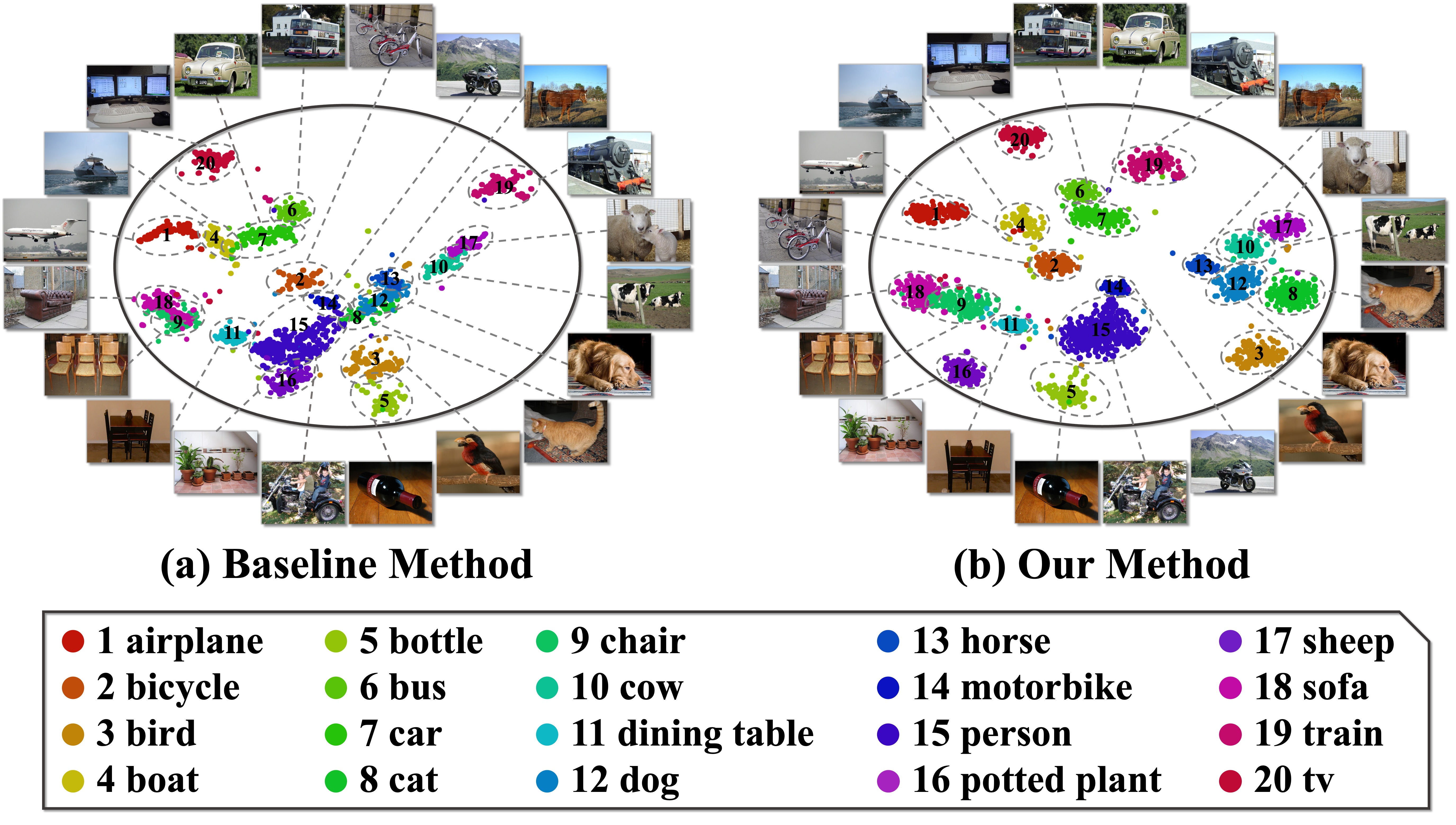}
\caption{ T-SNE results for the feature distribution of all unseen classes in the four folds (each contains 5 classes) on PASCAL-$5^i$.} \label{fig:9}
\end{figure}

\begin{table}[t]
\setlength{\abovecaptionskip}{1pt}
\setlength{\belowcaptionskip}{1pt}
\caption{Ablation studies of initialization scheme of FG-prompts, which are conducted under case (b) in Table \ref{tab:8}. ``SBERT" is a pre-trained language model and ``CLIP" is a pre-trained visual language model.} \label{tab:9}
\renewcommand\arraystretch{1.3}
\centering
\resizebox{0.82\linewidth}{!}{
\begin{tabular}{c|c|cc} 
\toprule
{\#} &Initialization Scheme & PASCAL-5$^i$ & COCO-20$^i$   \\ 
\hline
(a) &Random Initial        & 67.34     & 42.07         \\ 

(b) &Support Average Token  & 67.85     & 42.91        \\ 

(c) &SBERT \cite{reimers2019sentence}                 & \underline{68.03}    & \underline{43.28}        \\ 

(d) &CLIP \cite{radford2021learning}                 & \textbf{68.26}     & \textbf{43.75}        \\
\bottomrule
\end{tabular}}
\end{table}

\begin{table}[t]
\setlength{\abovecaptionskip}{1pt}
\setlength{\belowcaptionskip}{1pt}
\caption{ Ablation studies of Semantic Prompt Transfer (SPT). ``Support" and ``Query" indicate the semantic migration from support images to prompts and from the query image to prompts, respectively. ``$\mathcal{F}_G$'' indicates the Gaussian suppression.} \label{tab:10}
\renewcommand\arraystretch{1.3}
\centering
\resizebox{0.75\linewidth}{!}{
\begin{tabular}{c|ccc|cc} 
\toprule
{\#} &Support & Query & $\mathcal{F}_G$ & mIoU                      &  FB-IoU                        \\ 
\hline
(a) &\ding{51}       & \ding{55}      & \ding{51}                 & 71.07                 & \underline{81.4}                     \\
(b) &\ding{55}       & \ding{51}     & \ding{51}                 & 71.15                 & 81.1                     \\
(c) &\ding{51}       & \ding{51}     & \ding{55}                  & \underline{71.37}                 & 81.25                    \\
(d) &\ding{51}       & \ding{51}     & \ding{51}                 & \textbf{71.66}        & \textbf{81.59}           \\
\bottomrule
\end{tabular}}
\end{table}

\subsection{Ablation Studies} \label{Ablation Studies}
This section conducts a series of ablation studies to analyze the role of each component. Unless otherwise noted, the experiments are conducted on PASCAL-5${^i}$ benchmark with the backbone of DeiT-B/16 under the 1-shot setting.

\subsubsection{Component Analysis}
Table \ref{tab:8} demonstrates the impact of different components on performance and the following phenomena can be observed: (1) Introducing class-specific FG-prompts (FG) (case (b)) can achieve effective performance gains compared to baseline, e.g., 0.96\% mIoU on PASCAL-5$^i$ and 1.61\% on COCO-20$^i$, which demonstrates the effectiveness of the dynamic class-aware prompting paradigm for FSS. (2) After incorporating the Semantic Prompt Transfer (SPT), 2.27\% and 3.44\% mIoU gains on PASCAL-5$^i$ and COCO-20$^i$ (case (c)) can be achieved. These improvements highlight that extracting class-specific clues from support and query images significantly enhances the class-awareness, enabling the encoder to precisely focus on the objects. (3) Combining with the Part Mask Generator (PMG) to mine fine-grained part-level prompts achieves the performance gains of 3.38\% mIoU on PASCAL-5$^i$ and 4.78\% mIoU on COCO-20$^i$ (case (e)), which suggests that mining fine-grained part-level semantics further enable the role of prompts for more accurate segmentation. (4) When additionally introducing BG-prompts (BG) and enriching their description with SPT can achieve 4.36{\%} and 5.91\% mIoU gains (case (f)), illustrating the importance of background semantics for segmentation.

Fig.\ref{fig:8} illustrates the segmentation results for different variants of our method. Remarkably, compared to the baseline (3$^{rd}$ column), introducing FG-prompts better focuses the target region, in spite of interference from other irrelevant classes, e.g., the "person" next to the "bottle" (3$^{rd}$ row). SPT refines the description of the prompts to achieve more precise segmentation results (5$^{th}$ column), e.g., the leaves of the "potted plant" (2$^{nd}$ row). When further adding BG-prompts (6$^{th}$ column), it is possible to differentiate excellently between the target and background, e.g., the "person" on the motorcycle (1$^{st}$ row). The segmentation results in the last column show that PMG is committed to mining the part-level local semantics to derive more precise target regions and less false activation. 

In addition, we visualize the feature distribution of all unseen classes in the four folds on PASCAL-$5^i$ during the testing phase (see Fig.\ref{fig:9}). The class prototypes generated by performing MAP on the support features are projected into 2D space using t-SNE. It can be found that our method captures more discriminative unseen-class semantics, with more compact class representations and stronger inter-class variations. The above results explicitly demonstrate that PAT can drive the encoder to focus on different classes in a dynamically class-aware prompting manner, yielding category-representative features.

\begin{figure}[t]
\centering
\setlength{\abovecaptionskip}{1pt}
\setlength{\belowcaptionskip}{1pt}
\includegraphics[width=0.85\linewidth]{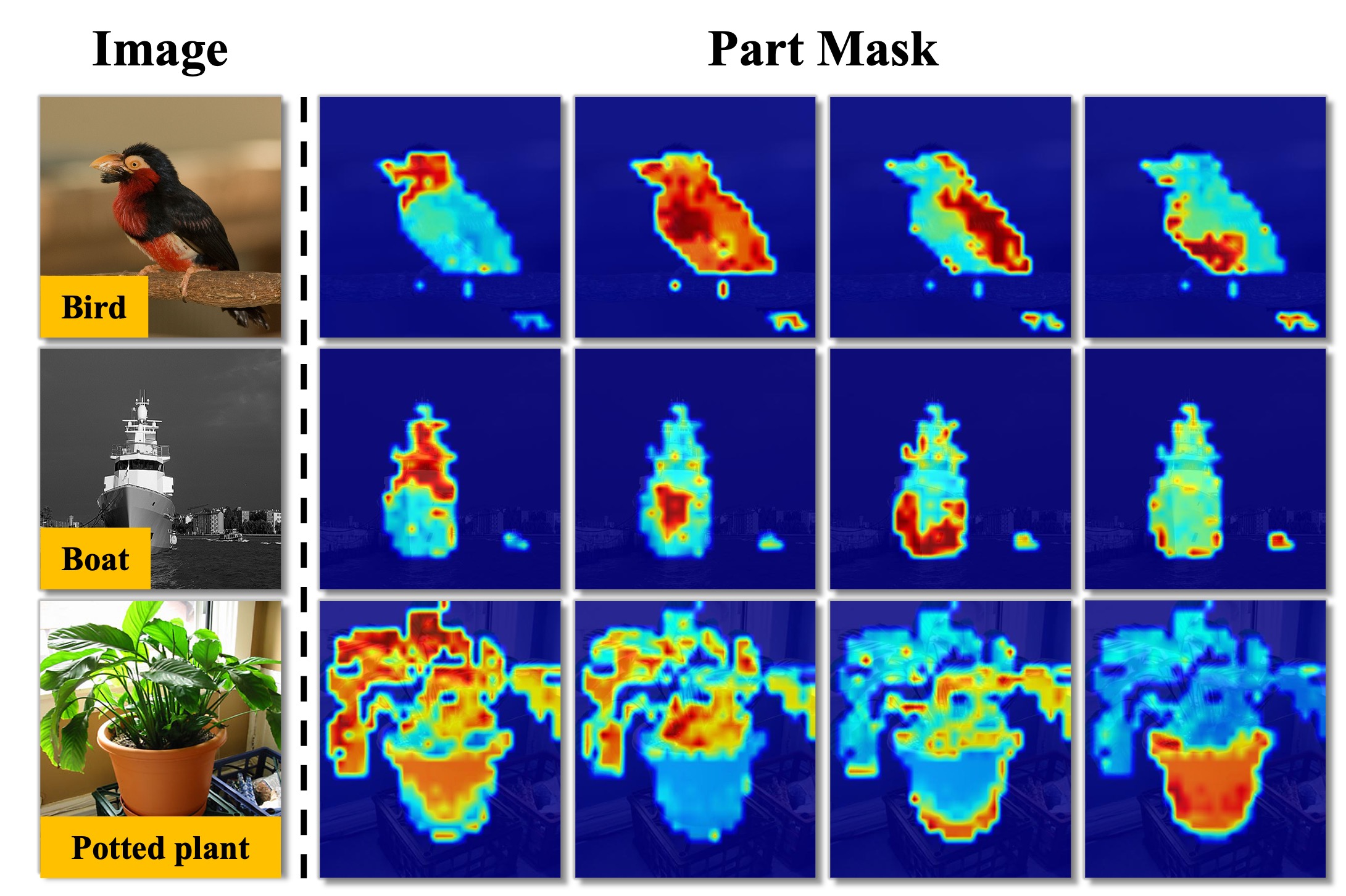}
\caption{ Visualization results for part masks in PMG (take four local masks as an example). It can be observed that for different tasks, different part masks adaptively focus on different part-level local regions.}
\label{fig:10}
\end{figure}

\begin{figure}[t]
\centering
\setlength{\abovecaptionskip}{1pt}
\setlength{\belowcaptionskip}{1pt}
\includegraphics[width=0.8\linewidth]{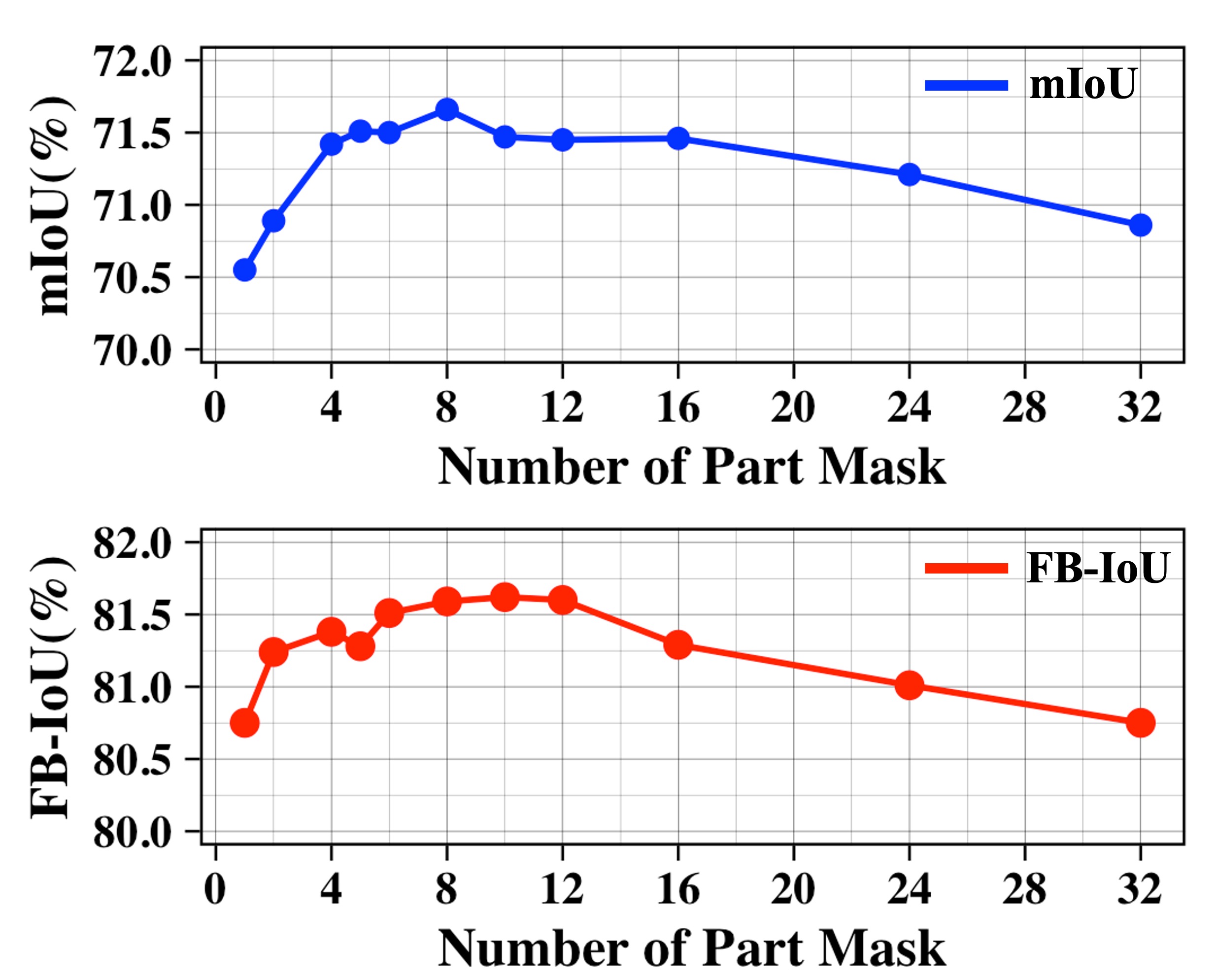}
\caption{ Ablation studies of the number of part masks on PASCAL-5$^i$ dataset in terms of mIoU and FB-IoU. Best mIoU accuracy is achieved when the number of part masks (i.e., $N_p$) is set to 8.}
\label{fig:11}
\end{figure}

\begin{table}[t]
\setlength{\abovecaptionskip}{1pt}
\setlength{\belowcaptionskip}{1pt}
\caption{ Ablation studies of the number of Prompt Enhancements. Given the deep features are more category-specific, we perform the Prompt Enhancement in the last $L$ blocks of the feature encoder.} \label{tab:13}
\renewcommand\arraystretch{1.3}
\centering
\resizebox{0.73\linewidth}{!}{
\begin{tabular}{c|ccccccc} 
\toprule
Layers & 1     & 2     & 3     & 5     & 7         \\ 
\hline
mIoU & 71.26  & 71.48 &71.66 &71.62 & 71.79   \\
FB-IoU & 81.19 & 81.42 & 81.59 & 81.67 & 81.71  \\
\bottomrule
\end{tabular}}
\end{table}

\begin{table}[t]
\setlength{\abovecaptionskip}{1pt}
\setlength{\belowcaptionskip}{1pt}
\caption{ Ablation studies of various encoder backbones on PASCAL-5$^i$. } \label{tab:11}
\renewcommand\arraystretch{1.3}
\centering
\resizebox{0.95\linewidth}{!}{
\begin{tabular}{c|c|cc|cc|c} 
\toprule
\multirow{2}{*}{\#} & \multirow{2}{*}{Backbone} & \multicolumn{2}{c|}{1-shot}     & \multicolumn{2}{c|}{5-shot}     & \multirow{2}{*}{\begin{tabular}[c]{@{}c@{}}Inference Speed\\(FPS)\end{tabular}}  \\ 
\cline{3-6}
&                           & mIoU           & FB-IoU         & mIoU           & FB-IoU         &                                                                                  \\ 
\hline
(a)                 & ViT-S/16                  & 64.41          & 77.02          & 70.99          & 81.85          & \textbf{21.58}                                                                   \\
(b)                 & ViT-B/16                  & 66.96          & 78.81          & 73.88          & 83.55          & 16.56                                                                            \\ 
\hline
(c)                 & DeiT-T/16                 & 62.78          & 76.39          & 68.76          & 80.88          & \underline{20.19}                                                                    \\
(d)                 & DeiT-S/16                 & \underline{68.65}  & \underline{79.85}  & \underline{73.94}  & \underline{84.01}  & 19.12                                                                            \\
(e)                 & DeiT-B/16                 & \textbf{71.66} & \textbf{81.59} & \textbf{78.01} & \textbf{86.43} & 14.59                                                                            \\
\bottomrule
\end{tabular}}
\end{table}

\begin{figure}[t]
\centering
\setlength{\abovecaptionskip}{1pt}
\setlength{\belowcaptionskip}{1pt}
\includegraphics[width=0.9\linewidth]{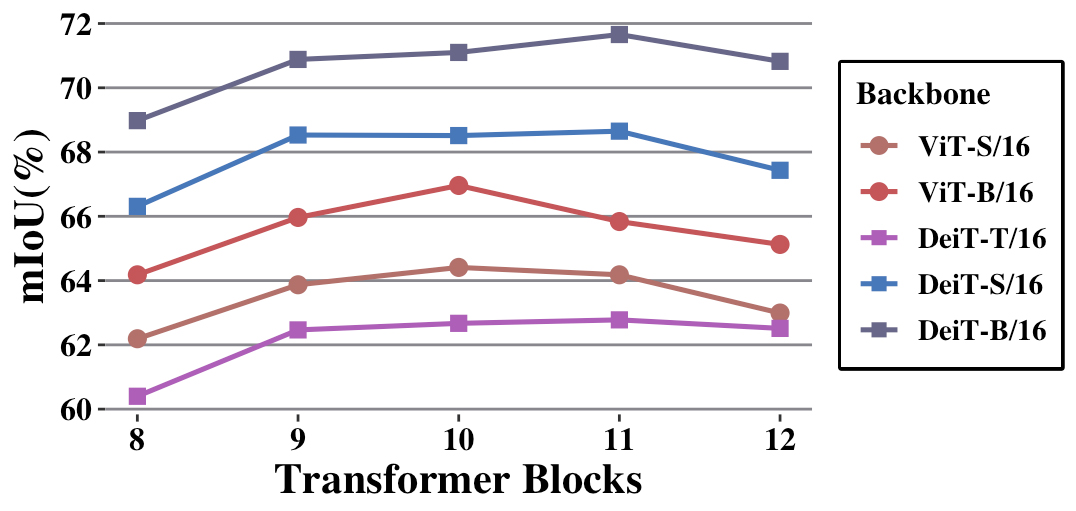}
\caption{ Ablation studies of the number of transformer blocks in different encoder backbones under the 1-shot setting.}
\label{fig:12}
\end{figure}

\subsubsection{Ablation study on Prompt Initialization}
We first examine the impact of the initialization scheme of FG-prompts on performance. Notably, this experiment is conducted under case (b) in Table \ref{tab:8} to highlight the effect of prompt initialization. From Table \ref{tab:9}, the following phenomena can be observed: (1) Tuning the encoder employing randomly initialized prompts yields no performance gains without SPT and PMG consistently enhancing their class-awareness. (2) Introducing additional category semantics as initial prompts can adjust the encoder to initially localize the target, boosting higher segmentation accuracy. (3) Linguistic information is more category-representative than the support average token. (4) Employing the linguistic information extracted by CLIP \cite{radford2021learning} as initial prompts has the most superior segmentation results.

\begin{table}[t]
\setlength{\abovecaptionskip}{1pt}
\setlength{\belowcaptionskip}{1pt}
\caption{Performance comparison of several FSS methods employing the transformer encoders on PASCAL-5$^i$ under the 1-shot setting. "$\dagger$" denote the results with the transformer encoders are from \cite{lang2024few} and \cite{zhang2022feature}. PAT$^\star$ indicates no additional linguistic information generated by CLIP is introduced for prompt initialization while PAT does the opposite.} \label{tab:12_0}
\renewcommand\arraystretch{1.3}
\centering
\resizebox{0.95\linewidth}{!}{
\begin{tabular}{r|cc|cc}
\toprule
\multirow{2}{*}{Method}                               & \multicolumn{2}{c|}{CNN Encoder} & \multicolumn{2}{c}{Transformer Encoder}                                       \\ 
\cline{2-5}
& ResNet-50 & ResNet-101   & ViT-B/16                         & DeiT-B/16                          \\ 
\hline
PMNet \cite{chen2024pixel}  & 65.40     & 68.10      & 62.83         & 64.72            \\
IPMT \cite{liu2022intermediate}   & 66.83     & 66.08      &  62.35        & 65.10           \\
PFENet$^\dagger$ \cite{tian2020prior} & 60.80     & 60.10      & 58.70    & 57.70      \\
CyCTR$^\dagger$  \cite{zhang2021few} & 64.20     & 64.30      & 60.10    & 61.00      \\ 
DCPNet$^\dagger$ \cite{lang2024few}  & 66.10     & 67.30      & 62.60    &  -           \\
\hline
PAT$^\star$ (Ours) & - & -          &66.26    &71.25      \\
{\cellcolor[rgb]{0.925,0.925,0.925}}\textbf{PAT (Ours)} &{\cellcolor[rgb]{0.925,0.925,0.925}} -         &{\cellcolor[rgb]{0.925,0.925,0.925}} -          &{\cellcolor[rgb]{0.925,0.925,0.925}} \textbf{66.96}    &{\cellcolor[rgb]{0.925,0.925,0.925}} \textbf{71.66}      \\
\bottomrule
\end{tabular}}
\end{table}

\begin{table}[t]
\setlength{\abovecaptionskip}{1pt}
\setlength{\belowcaptionskip}{1pt}
\caption{Ablation studies of PAT combined with existing decoder-based FSS methods on PASCAL-5$^i$. ``PAT$^\dagger$" denotes the dynamic class-aware encoder of the proposed PAT. ``$\Delta$" denotes the mIoU gains of utilizing PAT's class-aware encoder as the feature encoder of the previous FSS methods under the 5-shot setting.} \label{tab:12}
\renewcommand\arraystretch{1.3}
\centering
\resizebox{0.95\linewidth}{!}{
\begin{tabular}{c|cc|cc|c} 
\toprule
\multirow{2}{*}{Method} & \multicolumn{2}{c|}{1-shot} & \multicolumn{2}{c|}{5-shot} & \multirow{2}{*}{$\Delta$}  \\ 
\cline{2-5}
& mIoU  & FB-IoU              & mIoU  & FB-IoU              & \multicolumn{1}{l}{}                           \\ 
\hline
PFENet~\cite{tian2020prior}                 & 60.73 & 73.30               & 61.88 & 73.90               & \multirow{2}{*}{+8.46}                         \\
PFENet w/ PAT$^\dagger$ & 68.11 & 78.62               & 70.34 & 81.55               &                                                \\ 
\hline
IPMT~\cite{liu2022intermediate}                   & 66.83 & 77.10               & 68.20 & 81.40               & \multirow{2}{*}{+6.03}                         \\
IPMT w/ PAT$^\dagger$   & 70.10 & 78.95               & 74.23 & 83.37               &                                                \\ 
\hline
SSPNet~\cite{fan2022self}                 & 63.98 & -                   & 72.48 & -                   & \multirow{2}{*}{+6.88}                         \\
SSPNet w/ PAT$^\dagger$ & 73.19 & 81.81               & 79.36 & 87.70               &                                                \\ 
\hline
PAT                     & 71.66 & 81.59               & 78.01 & 86.43               & -                                              \\
\bottomrule
\end{tabular}}
\end{table}

\subsubsection{Ablation study on Prompt Enhancement}
\textbf{1) Part Mask Generator:} As stated in Sec.\ref{Part Mask Generator}, the Part Mask Generator (PMG) is designed to adaptively generate multiple part-level local masks to work with SPT to assign different part-level semantics for prompts. First, we qualitatively evaluate the visualization of part masks (see Fig.\ref{fig:10}). As can be noticed, the target object can be clearly divided into different complementary part-level local regions, e.g., the "bird" is divided into head, breast, back, and tail in the first row, which further demonstrates that the proposed PMG can adaptively generate different part-level masks.

Then, we examine the effect of various numbers of part masks on performance. As shown in Fig.\ref{fig:11}, mIoU accuracy increases as the number of part masks (i.e., $N_p$) rises from 1 to 8, because more part masks can mine more different part-level local semantics to refine the target description in the prompts. When continuing to increase, the mIoU accuracy decreases instead, especially with an increase to 32, where the mIoU falls below 71{\%}. This is because excessive part masks may result in the target object not being clearly divided into different interpretable local regions, thus creating redundancy and even noise from semantic fine-grained mismatches. A similar phenomenon is observed for FB-IoU.

\noindent \textbf{2) Semantic Prompt Transfer:} 

Table \ref{tab:10} demonstrates the effect of different variants in SPT on performance. The following phenomena can be derived: (1) Merely transferring either the support target semantics or the query target semantics to the prompts produces varying degrees of performance degradation, which is a good indication that for segmenting the query image, both accurate target semantics in support images and target semantics in query image itself are essential. (2) Gaussian suppression $\mathcal{F}_G$ in SPT allows for better aggregation of global semantics of specific regions to prompts by adjusting the attention distribution, thus yielding higher segmentation accuracy.

\noindent \textbf{3) The number of Prompt Enhancements:} 
Then, we analyze the effect of the number of Prompt Enhancements (i.e., the number of alternations for prompting and transferring) on performance, as illustrated in Table \ref{tab:13}. During the experiments, considering that deep features are more category-specific, we perform the Prompt Enhancement in the last $L$ blocks of the transformer encoder to assign richer category semantics to prompts. Overall, higher segmentation accuracy can be achieved by performing more semantic migrations, especially reaching a 71.79{\%} mIoU accuracy when performing 7 times (i.e., $L=7$). Considering the balance between efficiency and performance, we believe that 3 times (i.e., $L=3$) would be the appropriate choice.

\subsubsection{Ablation study on Backbone Setup}
This section examines the impact of various encoder backbones and the number of blocks on performance. First, according to Table \ref{tab:11} the following phenomenon can be observed: (1) Using the backbone of DeiT-B/16 has the best segmentation accuracy under all settings, while using the backbone of DeiT-S/16 achieves the second-best accuracy, which is even superior to the larger ViT-B/16. (2) Despite the poor segmentation performance achieved by using the smaller ViT-S/16 (88M) or DeiT-T/16 (23M) compared to other backbones, they have competitive inference speeds. Even so, their performance is comparable to some ResNet101-based (179M) FSS methods, eg., PFENet\cite{tian2020prior} (60.05{\%}) and CyCTR\cite{zhang2021few} (63.73{\%}).

Next, Fig.\ref{fig:12} demonstrates the effect of different numbers of blocks in various encoder backbones. It can be found that more blocks won't necessarily work better. For example, the two ViT variants work best with 10 blocks while the three DeiT variants work best with 11 blocks. Such a phenomenon is consistent with the fact that CNN-based FSS methods \cite{lang2022learning,liu2022intermediate} prefer mid-level features over high-level features.

\subsubsection{Comparison employing the Consistent Encoder}

To construct a fair comparison, we replace the CNN encoders of the competitive FSS methods PMNet \cite{chen2024pixel} and IPMT \cite{liu2022intermediate} with ViT-B/16 and DeiT-B/16 transformer encoders. In addition, for PAT, instead of introducing additional linguistic information generated by CLIP, we directly utilize the support average tokens as initial prompts. The performance results are illustrated in Table \ref{tab:12_0}. We can observe the following phenomenons: (1) With the consistent feature encoder setup, the proposed PAT achieves the best segmentation performance. For example, with the encoder of ViT-B/16, the PAT (without linguistic information) obtains 66.26\% mIoU, outperforming PMNet and IPMT by 3.43\% and 3.91\%, respectively. (2) Blindly employing the transformer to extract features may not yield the expected performance gains. Similar experimental findings were reported by DCPNet \cite{lang2024few}. This indicates that the transformer structure may not be applicable to the universal FSS structure with the frozen encoder and learnable decoders. (3) The performance gain of introducing additional linguistic information (i.e., 0.7\% mIoU gain with ViT) similarly demonstrates that cross-modal linguistic information facilitates more robust prompts for generating class-aware features.

\subsubsection{Combination of PAT and other FSS methods}

Since PAT aims to build a dynamic class-aware encoder for FSS, a natural exploration is whether performance can be consistently increased when PAT meets the previous decoder-based FSS methods (i.e., complex decoders). The segmentation results in Table \ref{tab:12} reveal the following phenomena: (1) Utilizing more complex decoders may not necessarily outperform the simple similarity computation. Possible reasons include: a) PAT already generates class-aware features focused on target classes, so reprocessing these features with complex decoders may have little effect and could even reduce accuracy. b) Directly utilizing features extracted from the transformer structure (i.e., PAT) for the CNN structure (i.e., PFENet and IPMT) may impact segmentation performance as ViT already constructs global associations through self-attention, making further local smoothing via convolutional layers potentially redundant. (2) Equipped with the proposed dynamic class-aware encoder, these FSS methods yield significant performance gains. Although PFENet and IPMT with the proposed class-aware encoder are not as effective as PAT itself, they perform significantly better compared to their original versions. The above gains indicate that compared to the frozen pre-trained encoder employed in previous FSS, the proposed dynamic class-aware encoder can generate class-aware features for different novel classes flexibly and is well-compatible with decoder-based methods.

\subsection{Extended Experiments} \label{Extended Experiments}
For further exploring the extensibility and flexibility of our PAT, we conduct a series of extension experiments in this section, including Cross-domain Few-shot Segmentation (see Sec.\ref{Cross-domain Few-shot Segmentation}), Weak-label Few-shot Segmentation (see Sec.\ref{Weak-label Few-shot Segmentation}), and Zero-shot Segmentation (see Sec.\ref{Zero-shot Segmentation}).

\subsubsection{Cross-domain Few-shot Segmentation} \label{Cross-domain Few-shot Segmentation}
To explore the generalization of FSS to unseen domains (datasets) in real-world situations, we follow the CWT\cite{lu2021simpler} and PATNet\cite{lei2022cross} setups to evaluate our method in cross-domain scenarios comprehensively. Specifically, the model trained in the meta-training phase is employed directly to the unseen domains for evaluation without any fine-tuning. Notably, the categories that appear in the source domain are removed from the unseen domain. First, we perform a cross-domain evaluation between PASCAL-5$^i$ and COCO-20$^i$. As indicated in Table \ref{tab:6}, our PAT beats other methods by a considerable margin under both cross-domain settings, establishing the new state-of-the-arts. To further explore the segmentation performance under larger domain shifts, we select three more complex datasets for evaluation, FSS-1000\cite{li2020fss}, Chest X-ray\cite{jaeger2014two}, and DeepGlobe\cite{demir2018DeepGlobe}, which cover 1,000 classes of daily objects, X-ray imagery, and six classes of satellite imagery, respectively. Following PATNet\cite{lei2022cross}, the proposed method is trained on PASCAL VOC and evaluated on these three datasets, with the mIoU results presented in Table \ref{tab:7}. Our method reaches top performance in all cases, especially yielding 15.84{\%} (1-shot) and 19.63{\%} (5-shot) mIoU gains on Chest X-ray. Furthermore, we visualize the cross-domain results on these three datasets, as illustrated in Fig.\ref{fig:13}. The above results strongly demonstrate that the proposed PAT can handle various cross-domain scenarios well and exhibits powerful robustness. 

\begin{table}[t]
\setlength{\abovecaptionskip}{1pt}
\setlength{\belowcaptionskip}{1pt}
\caption{ Extended experiments of Cross-domain Few-shot Segmentation between PASCAL-5$^i$ and COCO-20$^i$. ``-" denotes that the results are not recorded. Some comparison results are from \cite{lu2021simpler}.} \label{tab:6}
\renewcommand\arraystretch{1.3}
\centering
\resizebox{0.95\linewidth}{!}{
\begin{tabular}{r|c|cc|cc} 
\toprule
\multirow{2}{*}{Method}        & \multirow{2}{*}{Backbone} & \multicolumn{2}{c|}{COCO$\rightarrow$PASCAL} & \multicolumn{2}{c}{PASCAL$\rightarrow$COCO}  \\ 
\cline{3-6}
                   &                           & 1-shot         & 5-shot          & 1-shot         & 5-shot          \\ 
\hline
RPMMs \cite{yang2020prototype}                         & \multirow{4}{*}{ResNet-50} & 49.60          & 53.80           & 36.38          & 40.95           \\
CWT \cite{lu2021simpler}                           &                           & 59.45          & 66.50           & 39.15          & 45.75           \\
PFENet \cite{tian2020prior}                           &                           & 61.10          & 63.40           & -          & -           \\
RePRI \cite{boudiaf2021few}                        &                           & 63.20          & 67.70           & -              & -               \\ 

\hline
HSNet \cite{min2021hypercorrelation}                          &  \multirow{1}{*}{ResNet-101}                 & 64.10          & 70.30           & -              & -               \\

\hline
\rowcolor[rgb]{0.925,0.925,0.925} {\cellcolor[rgb]{0.925,0.925,0.925}}                                & \textbf{ViT-B/16}         & \textbf{66.86} & \textbf{75.29}                  & \textbf{40.78} & \textbf{47.07}                  \\
\rowcolor[rgb]{0.925,0.925,0.925} \multirow{-2}{*}{{\cellcolor[rgb]{0.925,0.925,0.925}}\textbf{Ours}} & \textbf{DeiT-B/16}        & \textbf{70.58} & \textbf{78.59}                  & \textbf{43.63} & \textbf{52.31}                  \\
\bottomrule
\end{tabular}}
\end{table}

\begin{figure*}[t]
\centering
\setlength{\abovecaptionskip}{1pt}
\setlength{\belowcaptionskip}{1pt}
\includegraphics[width=0.75\textwidth]{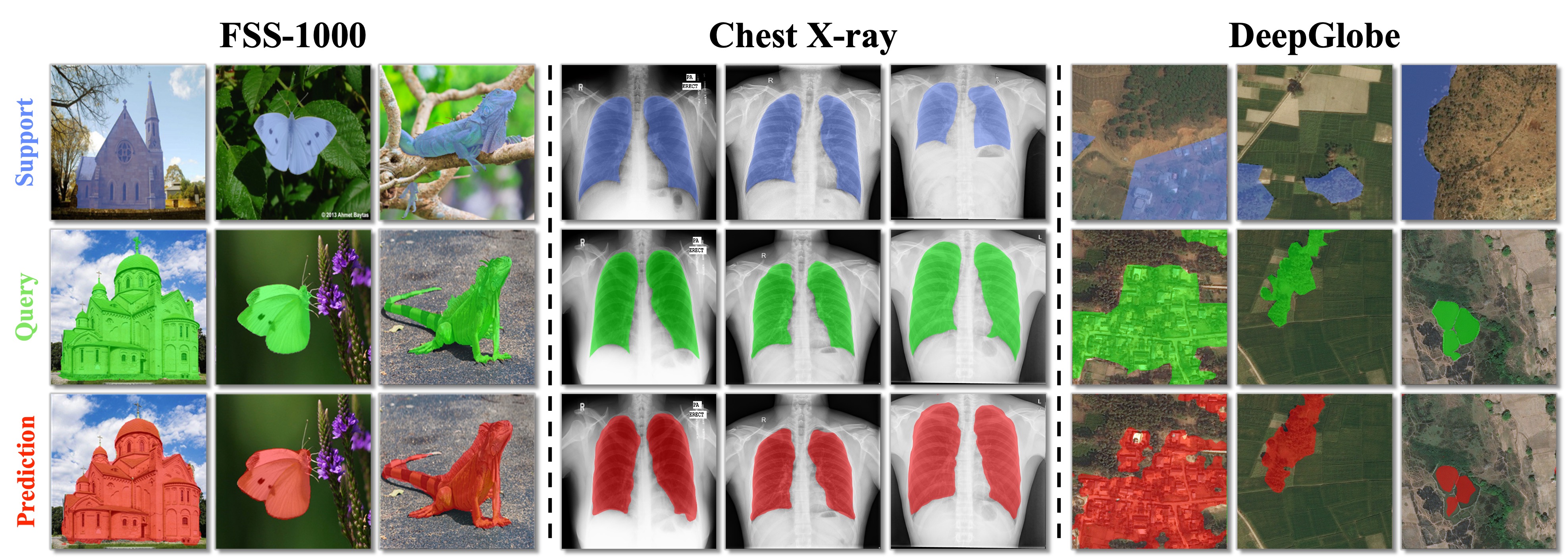} 
\caption{ Qualitative visualization results for Cross-domain Few-shot Segmentation under the 1-shot setting, where the model is trained on PASCAL VOC dataset and evaluated on three datasets with larger domain shifts, including FSS-1000, Chest X-ray, and DeepGlobe.}
\label{fig:13}
\end{figure*}

\begin{table*}[t]
\setlength{\abovecaptionskip}{1pt}
\setlength{\belowcaptionskip}{1pt}
\caption{ Extended experiments of Cross-domain FSS on three datasets with larger domain shifts. Notably, the proposed model is trained on PASCAL VOC and evaluated on these three datasets. Some comparison results are from \cite{lei2022cross}.} \label{tab:7}
\renewcommand\arraystretch{1.25}
\centering
\resizebox{0.7\linewidth}{!}{
\begin{tabular}{r|cc|cc|cc|cc} 
\toprule
\multirow{2}{*}{Method} & \multicolumn{2}{c|}{FSS-1000 (CV)}    & \multicolumn{2}{c|}{Chest X-ray (Medical)} & \multicolumn{2}{c|}{DeepGlobe (RS)} & \multicolumn{2}{c}{Average}   \\ 
\cline{2-9}
                        & 1-shot         & 5-shot         & 1-shot         & 5-shot          & 1-shot         & 5-shot  & 1-shot         & 5-shot  \\ 
\hline
PFENet \cite{tian2020prior}                 & 70.87          & 70.52          & 27.22          & 27.57           & 16.88          & 18.01        &38.32  &38.70   \\
RPMMs \cite{yang2020prototype}                  & 65.12          & 67.06          & 30.11          & 30.82           & 12.99          & 13.47     &36.07 &37.12      \\
RePRI \cite{boudiaf2021few}                  & 70.96          & 74.23          & 65.08          & 65.48           & 25.03          & 27.41     &53.69 &55.71      \\
HSNet \cite{min2021hypercorrelation}                   & 77.53          & 80.99          & 51.88          & 54.36           & 29.65    & 35.08  &53.02 &56.81         \\
PATNet \cite{lei2022cross}                  & 78.59          & 81.23          & 66.61          & 70.20           & 37.89          & 42.97   &61.03 &64.80       \\ 
\hline \rowcolor{gray!15}
\textbf{Ours}           & \textbf{81.15} & \textbf{83.44} & \textbf{82.45} & \textbf{89.83}  & \textbf{38.63} & \textbf{43.06} & \textbf{67.41} & \textbf{72.11} \\
\bottomrule
\end{tabular}}
\end{table*}

\begin{table}[t]
\setlength{\abovecaptionskip}{1pt}
\setlength{\belowcaptionskip}{1pt}
\caption{ Extended Experiments of Weak-label FSS on PASCAL-5$^i$. ``Dense'' denotes the densely labeled mask. ``Scribble'' and ``Bounding box'' denote the scribble label and bounding box label, respectively.} \label{tab:3}
\renewcommand\arraystretch{1.3}
\centering
\resizebox{0.95\linewidth}{!}{
\begin{tabular}{cr|cc|cc} 
\toprule
\multirow{2}{*}{Annotation}   & \multirow{2}{*}{Method}                           & \multicolumn{2}{c|}{1-shot}                                                                             & \multicolumn{2}{c}{5-shot}                                                                               \\ 
\cline{3-6}
                              &                                                   & mIoU                                               & FB-IoU                                             & mIoU                                               & FB-IoU                                              \\ 
\hline
\multirow{2}{*}{Dense}        & Baseline                                          & 67.30                                              & 77.56                                              & 75.50                                              & 84.76                                               \\
                              & {\cellcolor[rgb]{0.925,0.925,0.925}}\textbf{Ours} & {\cellcolor[rgb]{0.925,0.925,0.925}}\textbf{71.66} & {\cellcolor[rgb]{0.925,0.925,0.925}}\textbf{81.59} & {\cellcolor[rgb]{0.925,0.925,0.925}}\textbf{78.01} & {\cellcolor[rgb]{0.925,0.925,0.925}}\textbf{86.43}  \\ 
\hline
\multirow{2}{*}{Scribble}     

& Baseline                                          & 65.85                                              & 76.22                                              & 73.26                                              & 83.05                                               \\
                              & {\cellcolor[rgb]{0.925,0.925,0.925}}\textbf{Ours} & {\cellcolor[rgb]{0.925,0.925,0.925}}\textbf{69.08} & {\cellcolor[rgb]{0.925,0.925,0.925}}\textbf{77.92} & {\cellcolor[rgb]{0.925,0.925,0.925}}\textbf{75.13} & {\cellcolor[rgb]{0.925,0.925,0.925}}\textbf{84.19}  \\ 
\hline
\multirow{2}{*}{Bounding box} 

& Baseline                                          & 58.96                                              & 68.31                                              & 67.56                                              & 76.89                                               \\
                              & {\cellcolor[rgb]{0.925,0.925,0.925}}\textbf{Ours} & {\cellcolor[rgb]{0.925,0.925,0.925}}\textbf{64.83} & {\cellcolor[rgb]{0.925,0.925,0.925}}\textbf{73.92} & {\cellcolor[rgb]{0.925,0.925,0.925}}\textbf{69.15} & {\cellcolor[rgb]{0.925,0.925,0.925}}\textbf{79.37}  \\
\bottomrule
\end{tabular}}
\end{table}

\begin{figure}[t]
\centering
\setlength{\abovecaptionskip}{1pt}
\setlength{\belowcaptionskip}{1pt}
\includegraphics[width=0.92\linewidth]{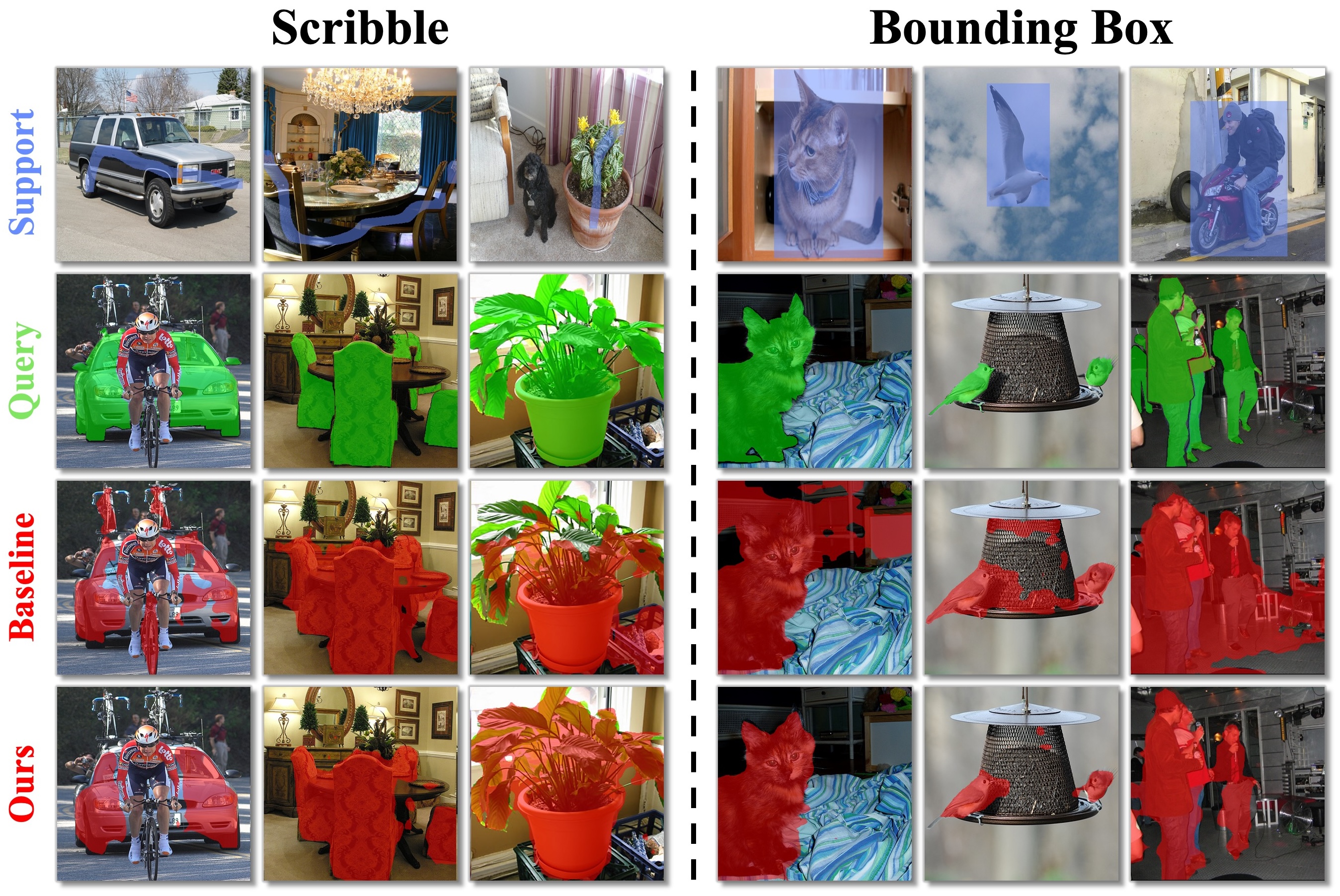} 
\caption{ Qualitative visualization results for Weak-label Few-shot Segmentation under the 1-shot setting.} 
\label{fig:14}
\end{figure}

\begin{table*}[htb]
\setlength{\abovecaptionskip}{1pt}
\setlength{\belowcaptionskip}{1pt}
\caption{ Performance comparison with state-of-the-art ZSS methods under the Zero-shot setting on PASCAL-5$^i$ and COCO-20$^i$. `$\ast $' denotes the variant of our method under the Zero-shot setting. } \label{tab:4}
\renewcommand\arraystretch{1.3}
\centering
\resizebox{0.95\linewidth}{!}{
\begin{tabular}{r|c|cccc|cc|cccc|cc} 
\toprule
\multirow{2}{*}{Method}                                                                               & \multirow{2}{*}{Backbone} & \multicolumn{6}{c|}{PASCAL-$5^i$}                                                                   & \multicolumn{6}{c}{COCO-$20^i$}                                                                      \\ 
\cline{3-14}
&                           & Fold-0         & Fold-1         & Fold-2         & Fold-3         & Mean           & FB-IoU         & Fold-0         & Fold-1         & Fold-2         & Fold-3         & Mean           & FB-IoU          \\ 
\hline
SPNet (CVPR'2019) \cite{xian2019semantic}                                                                                                 & ResNet-101                & 23.80          & 17.00          & 14.10          & 18.30          & 18.30          & 44.30          & -              & -              & -              & -              & -              & -               \\
ZS3Net (NeurIPS'2019) \cite{bucher2019zero}                                                                                                & ResNet-101                & 40.80          & 39.40          & 39.30          & 33.60          & 38.28          & 57.70          & 18.80          & 20.10          & 24.80          & 20.50          & 21.05          & 55.10           \\
\multirow{2}{*}{LSeg (ICLR'2022)\cite{li2022language}}                                                                                & ResNet-101                & 52.80          & 53.80          & 44.40          & 38.50          & 47.38          & 64.10          & 22.10          & 25.10          & 24.90          & 21.60          & 23.43          & 57.90           \\
& ViT-L/16                  & 61.30          & 63.60          & 43.10          & 41.00          & 52.25          & 67.00          & 28.10          & 27.50          & 30.00          & 23.20          & 27.20          & 59.90           \\
CLIPSeg (CVPR'2022) \cite{luddecke2022image}                                                                                              & ViT-B/16                  & 53.90          & 62.00          & 42.80          & 48.00          & 51.68          & 66.20          & 34.20          & 38.90          & 34.90          & 31.90          & 34.98          & 62.50           \\
UniBoost (arXiv'20223) \cite{sun2023uniboost}                                                                                              & ViT-L/16                  & 67.30          & 65.10          & 46.70          & 47.30          & 56.60          & 69.40          & 30.40          & 31.90          & 35.70          & 33.50          & 32.88          & 61.90           \\
\multirow{2}{*}{SAZS (CVPR'2023) \cite{liu2023delving}}                                                                                 & DRN                       & 57.30          & 60.30          & 58.40          & 45.90          & 55.48          & 66.40          & 34.20          & 36.50          & 34.60          & 35.60          & 35.23          & 58.40           \\
& ViT-L/16                  & 62.70          & 64.30          & 60.60          & 50.20          & 59.45          & 69.00          & 33.80          & 38.10          & 34.40          & 35.00          & 35.33          & 58.20           \\ 
\hline
\rowcolor[rgb]{0.925,0.925,0.925} {\cellcolor[rgb]{0.925,0.925,0.925}}                                & \textbf{ViT-B/16}         & \textbf{67.03} & \textbf{69.63} & \textbf{56.42} & \textbf{51.88} & \textbf{61.24} & \textbf{75.14} & \textbf{28.46} & \textbf{36.27} & \textbf{32.57} & \textbf{33.03} & \textbf{32.58} & \textbf{57.77}  \\
\rowcolor[rgb]{0.925,0.925,0.925} \multirow{-2}{*}{{\cellcolor[rgb]{0.925,0.925,0.925}}\textbf{Ours$^{\ast }$}} & \textbf{DeiT-B/16}        & \textbf{70.18} & \textbf{70.60} & \textbf{54.12} & \textbf{54.36} & \textbf{62.32} & \textbf{74.36} & \textbf{30.10} & \textbf{37.96} & \textbf{37.57} & \textbf{34.34} & \textbf{34.99} & \textbf{59.14}  \\
\bottomrule
\end{tabular}}
\end{table*}

\begin{figure}[t]
\centering
\setlength{\abovecaptionskip}{1pt}
\setlength{\belowcaptionskip}{1pt}
\includegraphics[width=0.85\linewidth]{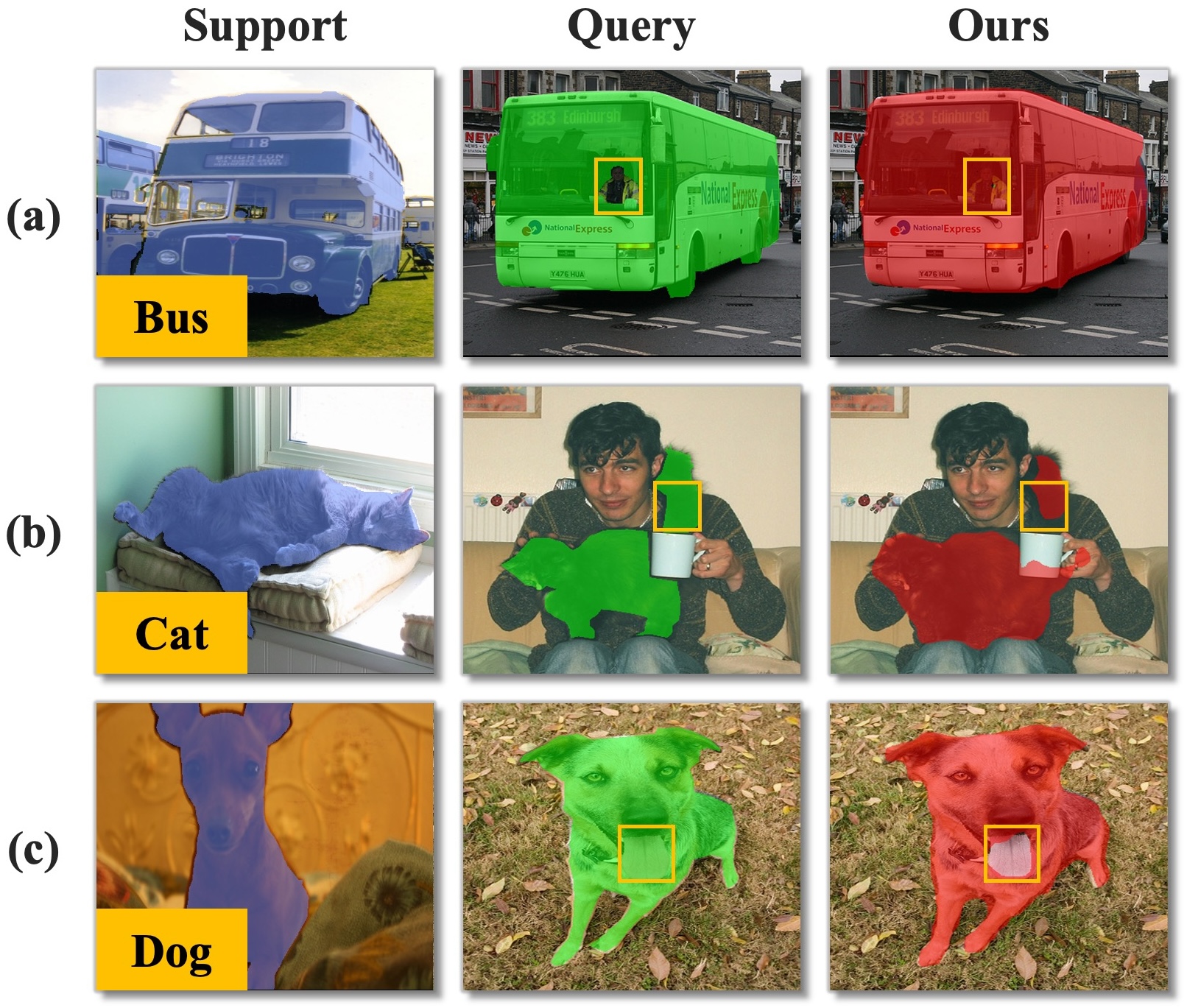}
\caption{ Several typical failure cases of our PAT. Note that the areas where the segmentation fails are highlighted with yellow boxes.}
\label{fig:16}
\end{figure}
\subsubsection{Weak-label Few-shot Segmentation}\label{Weak-label Few-shot Segmentation}
To alleviate the burden imposed by dense mask labeling in real-world applications, we further explore the effectiveness of FSS with weak-label samples. We employ two cheaper weak-label schemes for evaluation, i.e., scribble and bounding box, which are randomly produced from the original dense labeled masks (see \cite{wang2019panet} for more details). As can be seen from Table \ref{tab:3} and Fig.\ref{fig:14}, our method significantly beats the baseline in all cases, whatever the type of labeling. More interestingly, our method with scribble annotations achieves a 69.08{\%} mIoU accuracy, which is comparable to the accuracy of the best method HDMNet\cite{peng2023hierarchical} with dense annotations (i.e., 69.35{\%} mIoU). The above results indicate that our PAT is highly tolerant to the annotation quality and exhibits strong robustness both in the case of insufficient annotation information and excessive noise.

\subsubsection{Zero-shot Segmentation}\label{Zero-shot Segmentation}
Our PAT is also extended to the more challenging but realistic scenario, i.e. Zero-shot Segmentation (ZSS), which aims to segment unseen categories without any annotated sample. Even though our method is designed for the standard FSS task, it can be readily adapted to the ZSS task. Specifically, we still exploit CLIP to encode category names as initial FG prompts, which utilize SPT to continuously capture the target semantics of the query image. Notably, we remove the background prompts and focus only on the target foreground. For decoding, we utilize several convolutional layers to directly predict the segmentation result instead of the original Matching Head. 

Table \ref{tab:4} demonstrates the comparison of our method with other ZSS methods. As can be noticed, our variant enables competitive segmentation accuracy compared to other methods designed specifically for ZSS. In particular, our variant with the backbone of DeiT-B/16 outperforms the best method by 2.87{\%} mIoU and 4.96{\%} FB-IoU on PASCAL-5$^i$, setting a new state-of-the-art. Notably, our variant even outperforms several methods under the 5-shot setting, such as PFENet's 61.88{\%} mIoU in Table \ref{tab:1}. The above satisfactory results strongly prove that the proposed prompt-driven method can effectively focus on specific objects, exhibiting good robustness and extensibility.

\subsection{Failure case analysis}\label{Failure case analysis}
Sec.\ref{Comparison with the State of-the-Arts}-\ref{Extended Experiments} have shown that the proposed PAT accomplishes efficient segmentation. Yet, there are still several failure cases that require further attention. From Fig.\ref{fig:16}(a)-(b), false segmentation may occur in cases where the target and background overlap or have particularly similar textures and colors, e.g., the "person" and the "bus", the tail of the "cat" and the sweater. In addition, when significant appearance differences exist between target pixels, e.g., the tongue and body of the "dog" have significantly different color textures, the model may fail to focus on all target pixels effectively. We offer several thoughts that perhaps help to address these shortcomings: (1) Introducing more detailed text descriptions rather than category names to help segment specific categories. (2) Utilizing diffusion models to generate multiple support images to provide richer category semantics instead of one/few-shots.

\section{Conclusion}
In this paper, discarding the previous FSS practice of freezing the encoder for generalization to unseen classes, we mimic the visual perception pattern of human beings and introduce a novel dynamic class-aware prompting paradigm (PAT) to tune the encoder for focusing on specific objects in different FSS tasks. Our method sets new state-of-the-art (SOTA) on three popular FSS benchmarks. Surprisingly, when extended to more realistic scenarios including cross-domain, weak-label and even zero-shot, it yields equally satisfactory results. We hope that our work could offer fresh perspectives and inspire future research to focus on designing encoders for few-shot scenarios.


%



\ifCLASSOPTIONcompsoc
  \section*{Acknowledgments}
\else
  \section*{Acknowledgment}
\fi

This work was supported by the National Natural Science Foundation of China (NSFC) under Grant 62301538.

\bibliographystyle{ieeetr}
\bibliography{main}

\ifCLASSOPTIONcaptionsoff
  \newpage
\fi

\end{document}